\documentclass{article}

\usepackage{arxiv}
\usepackage{adjustbox}
\usepackage[utf8]{inputenc}
\usepackage[T1]{fontenc}
\usepackage{setspace}
\usepackage{graphicx}
\usepackage{epstopdf}
\usepackage[labelsep=period]{caption}  % added 14-04-2016 Markus Englich - Recommendation by Sebastian Brocks
\usepackage[british]{babel} 
\usepackage[hang]{footmisc}
\usepackage{multirow}
\usepackage{amsmath,amsfonts}
\usepackage{algorithmic}
\usepackage{algorithm}
\usepackage{array}
\usepackage[caption=false,font=footnotesize]{subfig}
\usepackage{textcomp}
\usepackage{stfloats}
\usepackage{url}
\usepackage{verbatim}
\usepackage{booktabs}
\usepackage[table,xcdraw]{xcolor}
\usepackage{enumitem}
\usepackage{tikz}
\usepackage[colorlinks=false,hidelinks]{hyperref}

\usepackage[authoryear]{natbib}
\providecommand{\abstract}[1]{\begin{abstract}#1\end{abstract}}
\providecommand{\keywords}[1]{\vspace{0.5em}\noindent\textbf{Keywords:} #1}

\begin{document}

\tolerance=999
\sloppy
\setlength{\emergencystretch}{2em}

\title{Pixel-Wise Multimodal Contrastive Learning for Remote Sensing Images}

\author{
Leandro Stival \\
Wageningen University \& Research \\
6708~PB Wageningen, The Netherlands \\
\texttt{leandro.stival@wur.nl} \\
Institute of Computing, University of Campinas (UNICAMP) \\
13083--852 Campinas, SP, Brazil \\
\texttt{leandro.stival@ic.unicamp.br} \\
\And
Ricardo da Silva Torres \\
Wageningen University \& Research \\
6708~PB Wageningen, The Netherlands \\
\texttt{ricardo.dasilvatorres@wur.nl} \\
\And
Helio Pedrini \\
Institute of Computing, University of Campinas (UNICAMP) \\
13083--852 Campinas, SP, Brazil \\
\texttt{helio@ic.unicamp.br} \\
}

\maketitle

\begin{abstract}
Satellites continuously generate massive volumes of data, particularly for Earth observation, including satellite image time series (SITS). However, most deep learning models are designed to process either entire images or complete time series sequences to extract meaningful features for downstream tasks. In this study, we propose a novel multimodal approach that leverages pixel-wise two-dimensional (2D) representations to encode visual property variations from SITS more effectively. Specifically, we generate recurrence plots from pixel-based vegetation index time series (NDVI, EVI, and SAVI) as an alternative to using raw pixel values, creating more informative representations. Additionally, we introduce PIxel-wise Multimodal Contrastive (PIMC), a new multimodal self-supervision approach that produces effective encoders based on two-dimensional pixel time series representations and remote sensing imagery (RSI). To validate our approach, we assess its performance on three downstream tasks: pixel-level forecasting and classification using the PASTIS dataset, and land cover classification on the EuroSAT dataset. Moreover, we compare our results to state-of-the-art (SOTA) methods on all downstream tasks. Our experimental results show that the use of 2D representations significantly enhances feature extraction from SITS, while contrastive learning improves the quality of representations for both pixel time series and RSI. These findings suggest that our multimodal method outperforms existing models in various Earth observation tasks, establishing it as a robust self-supervision framework for processing both SITS and RSI.
\end{abstract}


\section{Introduction}

% Como dados de sensioramento remonto sao improtantes
The use of remote sensing imagery (RSI) and multispectral remote sensing imagery (MSRSI) is crucial for Earth Observation (EO) tasks, such as forest monitoring~\citep{banskota2014forest,huete2012vegetation}, climate change analysis~\citep{diaz2024trends}, and crop estimation~\citep{trentin2024tree}. To effectively monitor changes and events on the Earth's surface, it is essential to utilize RSI and incorporate temporal information about these locations. In this context, satellite image time series (SITS) have proven to be highly effective for supporting these tasks~\citep{tarasiou2023vits}.

Nowadays, most of these images and time series are processed using machine learning models. However, obtaining high-quality labeled data remains a significant challenge in this field, as labeling SITS requires substantial time and effort from experts. Given the labor-intensive nature of this task, one of the most common approaches to training models on SITS and RSI is through self-supervised methods, such as contrastive learning~\citep{manas2021seasonal,mall2023change}.

% A informacao temporal dentro do SSL poderia ser melhor aproveitada
Most studies on using SSL for SITS concentrate on processing entire images to create robust feature representations. However, processing the entire image can often lead to higher computational costs and unnecessary overhead, especially in cases where only specific parts of the RSI are the focus of the analysis or monitoring.

% Os metodos focam em representacoes que nao a nivel de pixel, e que isso pode trazer ganhos
In this context, approaches that utilize only pixel information from SITS have shown promising results for SITS classification~\citep{garnot2020satellite}. However, these methods rely solely on pixel values to represent the time series, overlooking the potential of using features extracted from RSI that offer richer spatial context information (e.g., geometric and morphological metrics, vegetation indices, and texture patterns). In this study, we use vegetation indices to process SITS pixels, allowing us to utilize more detailed information to train effective encoders.

% Uso de 2D tem mais info do que 1D

Pixel-based SITS representations typically consist of one-dimensional (1D) data, where pixel values at each timestamp are stored and processed. This approach tends to overlook internal relationships between the values within the series~\citep{huamin2020reconstruction}. To enhance the representation of pixel data from SITS, we use the recurrence plot (RP)~\citep{eckmann1995recurrence} method, which transforms a 1D time series into a 2D representation. The choice of the recurrence plot is motivated by its successful use in classification and retrieval tasks, including SITS and near-surface image sequences~\citep{Dias2020GRSL,Menini2019GRSL,Faria2016PRLA,dias2020multirepresentational}. This not only makes temporal correlations among time samples explicit but also allows us to explore the use of machine learning models well established for computer vision tasks.
RP has also been successfully employed in time series analyses in multiple domains such as statistical analyses~\citep{da2024statistical}, magnetic resonance~\citep{aithal2024mci}, electroencephalogram (EEG)~\citep{goel2024automated}, and sports injury~\citep{ye2023novel}. Nevertheless, the aforementioned studies merely focus on using transfer learning methods using pre-trained computer vision models in supervised learning scenarios. In contrast, our approach employs 2D representations more generically, enabling the utilization of unlabeled datasets for computing representations of multiple modalities using the same input sample (remote sensing image).

% Utilizar indices de vegetacao em SITS de forma contrastiva
In this paper, we introduce Pixel-Wise Image Multimodal Contrastive (PIMC), a multimodal contrastive method for training machine learning models using 2D representations of the pixel information from SITS alongside RSI. To validate the superiority of encoders trained with the 2D representation, we benchmarked models trained in a supervised manner on 1D time series and compared their performance with the self-supervised PIMC.

% O que pretendemos responder
This paper addresses the following research questions:
\begin{itemize}

\item RQ1: How to create a multimodal framework that utilizes two-dimensional time series representations and MSRSI to train a machine learning model in a self-supervised multimodal contrastive way using vegetation index information from SITS?

\item RQ2: Are the machine learning models trained following the PIMC protocol effective in different downstream tasks (e.g., time series classification and forecasting, and image classification)?

\item RQ3: Would the PIMC approach yield results comparable to or better than state-of-the-art supervised and self-supervised models in downstream tasks?

\end{itemize}

In short, this work introduces a new multimodal self-supervised contrastive learning to train effective encoders and investigates their use in downstream tasks involving time series and RSI data only. To the best of our knowledge, this is the first study concerning the use of such modalities in self-supervision learning based on contrastive learning. 

% to produce 2D representations from SITS, in which the features relate to vegetation indices. The goal is to use these 2D representations to train two self-supervised encoders that are able to extract effective features from RSI images and the 2D representations of time series for EO-related downstream tasks.

The main contributions of this paper are:

\begin{itemize}
    \item A new contrastive self-supervised method to align the latent spaces from pixel-wise time series and remote sensing images.

    \item Demonstration of the effectiveness of encoders when trained using 2D representations of pixel-wise temporal information in SITS in various RSI-related downstream tasks.
    
    \item Demonstration of the effectiveness of trained encoders when compared with SOTA methods in different downstream tasks involving both modalities.
    
\end{itemize}

\section{Background and Related Work}

This section overviews background concepts and relevant studies to the topic under investigation in this paper.

\subsection{Remote Sensing Images}

The widespread use of remote sensing imaging enables the daily production of vast amounts of data from the Earth's surface. This abundance of data has fueled the growing application of machine learning models to a wide range of Earth observation (EO) tasks.

In examining the current literature, we can highlight several studies~\citep{Mall_2023_CVPR_CACo,manas2021seasonal,akiva2022self} that utilize self-supervision methods to create effective feature extractors for various tasks, such as land use classification, semantic segmentation, and change detection. Following this idea of domain generalization for EO applications, foundational models have also led to an increase in the number of publications proposing models that can be used or fine-tuned for common EO tasks~\citep{hong2024spectralgpt,Guo_2024_CVPR,xiong2024neural}.

Nevertheless, all of these methods for creating RSI representations share one common aspect: they utilize the entire image. However, since we are discussing RSI, it is important to recognize that each individual pixel represents a real-world region, typically spanning tens of meters. These regions often exhibit significant differences from their neighbors, particularly in areas with high levels of human intervention, such as cities and agricultural fields. In line with prevailing trends in the EO literature, this work employs unlabeled data to pre-train models in a self-supervised manner. However, unlike the conventional approach of using all pixels in the image to construct the SITS representation, we have chosen to analyze the data at the pixel level, resulting in a much more compact representation. This approach allows for a more focused examination of the nuances in local temporal changes.

\subsection{Self-Supervised Learning}
The training process employing self-supervised learning typically comprises two distinct categories of training methods: autoencoders (Section~\ref{subsubsec:auto_encoders}) and contrastive learning (Section~\ref{subsubsec:contrastive_learning}). This section introduces some of the most recent applications of SSL in time-series and remote sensing applications.

\subsubsection{Autoencoder-based Learning}
\label{subsubsec:auto_encoders}

In our study, we introduce a multimodal SSL scheme that accounts for pixel-wise information related to time series and local spatial patterns within SITS. Existing autoencoder-based SSL initiatives found in the literature often focus on learning from those modalities {\em in isolation}. Existing multimodal approaches often account for different types of remote-sensing images.

Examples of SOTA SSL for time series data include MOMENT~\citep{goswami2024moment} and Lag-llama~\citep{rasul2024lagllama}. MOMENT~\citep{goswami2024moment} is a model trained based on a masked-auto encoder scheme that can represent trend, scale, and frequency patterns. The training was conducted using the Time-series Pile dataset, composed of thousands of public datasets. Lag-llama~\citep{rasul2024lagllama} is another foundation model for the general time series. One novelty in its formulation refers to the use of a tokenization strategy, where the longer sequences are constructed using \emph{lags} (e.g, weeks, months, hours) to guide the model to learn how to represent different time granularity of the data in an effective way.

The utilization of SSL has also received significant attention within the RSI literature. A promising method was introduced by~\citet{XUE2025110959}. Their approach performs SSL for training an asymmetric encoder-decoder structure that utilizes cross-attention layers to facilitate the extraction of features from RSI data. The training process employs three distinct types of RSI: very high resolution (VHR), digital surface model (DSM), and hyperspectral images (HSI). The transformer encoder processes these features, and different decoders (depending on the type of images) then process them to reconstruct the original image. This process has demonstrated positive results in classification tasks related to land cover. To some extent, their method is multimodal as it considers multiple types of RSIs. Different from our initiative, their training process does not account for pixel-wise information or time series.

Another recent work was proposed by~\citet{qu2025self}. Their study was centered on the change detection task, wherein the utilization of images from different types (e.g., SAR images and optical images) enables the training of a transformer-based encoder to distinguish representations of two pixels over time, considering both types of images. The main contribution of their study lies in the integration of a Unified Mapping Unit (UMU) during the training phase. In this phase, four images are considered as input: two modalities (image types) for two different timestamps. Similar to our work, their study accounts for temporal variation. However, only two timestamps are considered. Another difference concerns the modalities considered in the training of the encoder. Finally, their study was validated on a single downstream task. %The training process employs a cross-entropy loss to preserve the similarity among representations from the same region across different modalities. Additionally, a fusion method is implemented to generate a single feature representation for all modalities.

\citet{sanchez2024self} also explored multimodality by employing two distinct types of images from the same region: LiDAR and hyperspectral. A random forest model is applied to select the optimal bands from the hyperspectral images, and later, SSL training is implemented through an enhanced nearest neighbor contrastive network. This training protocol utilizes a contrastive method to align different views of the same image. No temporal or pixel-wise information is considered in their study.%, but with changes to the augmentation typically presented in the MoCo for multimodality images. %As a result of these changes, the encoder presented better accuracy results when compared to the original NNCNet and other transformer-based encoders.

\citet{hou2025self} developed a classification model that demonstrates effective generalization capabilities in few-shot settings. The training process is structured into two distinct stages. Initially, a supervised learning process is employed to facilitate the extraction of features from RSIs. Subsequently, the SSL technique is implemented to refine these features and enhance the model's resilience in few-shot scenarios. The primary contribution of this approach lies in the integration of features from both the supervised and SSL training stages. Again, no temporal or pixel-wise properties are considered in the training process. %This integration serves to direct the construction of feature spaces through the utilization of a graph convolutional network (GCN), thereby facilitating the establishment of connections between representations based on the graph of sample similarity.

\citet{liu2024multi} conducted a comparative analysis, utilizing various contrastive SSL methodologies that were trained on the Sentinel2GlobalLULC dataset composed of multispectral and Synthetic-aperture radar (SAR) images. These encoders were then evaluated as feature extractors for land cover classification tasks. The findings indicated that employing these techniques as a pre-training process yielded superior outcomes in comparison to random initialization and pre-trained weights from natural image datasets (e.g., ImageNet), particularly when training with a reduced number of images. Unlike our approach, the method does not account for temporal and pixel-wise information. Furthermore, performed analysis concerned with only one downstream task.

%Descrever como nosso SSL multimodal can ser usado em diferentes modais independente
The literature presents various SSL methods that cater to different tasks, model architectures, and modalities. However, the current study is the first to apply SSL using both temporal pixel-wise information and remote sensing imagery to create two distinct encoders.

\subsubsection{Contrastive Learning}
\label{subsubsec:contrastive_learning}

Contrastive learning is a technique extensively explored in recent studies for self-supervised learning based on SITS. Relevant examples include SimCLR~\citep{chen2020simple} for general approaches, MoCo~\citep{he2020momentum} for RSI, and SeCo~\citep{manas2021seasonal} and CACo~\citep{mall2023change} for multispectral remote sensing images (MSRSI). Additionally, contrastive learning has also been applied to raw time series~\citep{zhang2024self}.

The contrastive learning method is an effective approach for bringing the representations of similar samples (positive pairs) closer together in the feature space while increasing the dissimilarity between other samples. This makes it a valuable technique for unlabeled datasets, where pairs can be generated through random alterations of the samples.
However, using augmented versions of samples is not the only way to implement contrastive methods. To enhance the amount of data and create more generalized representations, multimodal approaches can be employed in RSI applications. For instance, in the work described by~\citet{jain2022multimodal}, pairs of RSI from Sentinel-1 and Sentinel-2 satellites covering the same regions were used as positive pairs.

Since our approach emphasizes the 2D representation of SITS to generate a local representation of the time series, we integrate it with the concept of multimodal contrastive training. This concept is based on the classical learning of transferable visual models from natural language supervision (CLIP)~\citep{radford2021learning}. In this multimodal training involving text-image pairs, the objective is to train two encoders to converge toward similar feature spaces for both modalities.

In our study, instead of utilizing text and images as CLIP does, we employ the RSI and 2D representations of SITS vegetation indices at the pixel level. To the best of our knowledge, this represents an original application of the 2D representation of SITS at the pixel level in conjunction with RSI data, creating a robust feature space that can be utilized across various tasks involving both time series and RSI.

\subsection{Pixel-wise Approaches}

% Pixel wise e um metodo classifoc em computer vision
The process of pixel-by-pixel image processing has been a standard method of image analysis since its inception. Instead of processing the entire image, this approach allows for the examination of individual pixels, facilitating a more detailed and precise understanding of the image content.

% Quem ja usou e como usou
Many applications can benefit from this method of examining the data, such as (i) Soil Evaporative Efficiency (SEE)~\citep{SUN2023108063}, where optical and thermal data are used to estimate soil moisture, (ii) Super-Resolution in RSI~\citep{rs15123139}, where each pixel is processed to enhance the resolution of low-resolution bands, and (iii) Hyperspectral Image Classification~\citep{10146301,pedrini2016}, which employs different points within the same image for classification.
% Como esta atualmente na literatura
Additionally, authors have utilized pixel-wise approaches with attention mechanisms, such as the Pixel Attention Network~\citep{zhao2020efficient} and PiCANet~\citep{Liu_2018_CVPR}.

% Como pretendemos adiconar mais info na literatura sobre pixel wise ( o que e de novo)
The aforementioned methods introduced novel methods for information processing at the pixel level. However, there is currently no study addressing multimodal self-supervision learning based on two-dimensional pixel-wise time series representations and remote sensing images.

\subsection{Satellite Image Time Series}

Satellite image time series (SITS) have become a significant topic in recent literature. Typical applications involve predicting the future states of specific regions at various timestamps. 

%Usa SITS para classification de pixels, but nao fazer forecast
\citet{yan2024tsanet}, for example, introduced TSANet, a transformer-based supervised model that is trained on a dataset of SITS. The main objective of the model is to learn how to represent spatial-spectral-temporal features. %To demonstrate the efficacy of the proposed method, the authors conducted the classification of the pixels in the images and estimate the boundaries of each crop region. 
%The training process employed a combination of a segmentation and boundary loss function. However, the 
Their study demonstrates the use of temporal information for predicting crop types and their boundaries in RSI. Similar to our study, TSANet considered spatial and temporal information. However, the learning process is supervised and does not account for pixel-wise information.

Another SITS-based method was introduced by~\citet{dumeur2024self}. Their work presented UNet-BERT, a multimodal SSL methodology based on the BERT~\citep{devlin2018bert}. UNet-BERT is trained to generate feature representations based on spatial, spectral, and temporal data. The main contribution of this methodology is the U-BARN module, which integrates a patch embedding in a transformer block to produce features for all the modalities at once. These features are then employed by a shallow classifier to predict pixels in a semantic segmentation task similar to a masked autoencoder. Different from our approach, UNet-BERT requires all modalities for making inferences. Our method employs multimodality exclusively during training, yielding two independent encoders for each modality. Another distinction refers to the validation of UNet-BERT in a single downstream task (semantic segmentation).

$\alpha\text{-TIM}$~\citep{mohammadi2024few} is another method tailored to SITIS. The proposed formulation employs few-shot learning (FSL) techniques using a transformer-based method with a temporal pyramid layer to produce a crop mapping method. The main contribution of the method is the use of a Dirichlet distribution in the set of the few-shot query, while the model is previously trained in a supervised way. %The results obtained have presented a good pixel classification values for the classes that have fewer examples after do a fine tuning using the FSL techniques. 
$\alpha\text{-TIM}$ only employs temporal information in the training process. Another difference refers to the supervised training setting.

% Contra ponto (temos um encoder para cada modal, nao precisamos de labels, os modais podem ser alterados mantendo o mesmo protocolo de treino).
In light of the existing literature on SITS, we emphasize our methodology, PIMC, which exploits multimodality and temporal information at the pixel level while yielding independent results for RSI and temporal series. Furthermore, our approach can be readily adapted to utilize diverse data modalities (e.g., SAR images, superpixels, vegetation indices). This flexibility enhances the general applicability of our proposed SSL training protocol, allowing it to function effectively across different types of datasets, whether labeled or unlabeled.

\subsection{Time Series Representations}

% Formas de presentar time series
When examining the sequence of pixels in the same region of an RSI, this information can typically be represented as a time series, where each point in the series corresponds to the pixel value at a specific time. The representations of pixel values are typically defined as sequences of one-dimensional values, reflecting the inherent one-dimensional nature of the data. In the literature, various formulations exist to encode time series properties through transformations. One frequently used transformation in the time series domain involves converting univariate one-dimensional representations into two-dimensional representations.

% metodos que utilizam
The main methods available for this type of representation include Recurrence Plots (RP)~\citep{eckmann1995recurrence}, Gramian Summation Angular Fields (GASF), Gramian Difference Angular Fields (GADF)~\citep{wang2015imaging}, and Markov Transition Fields (MTF)~\citep{wang2015imaging}. These methods transform the data in ways that accentuate temporal patterns, highlight similarities or other intrinsic relations between points in a series, and reveal both the global and local behavior of the data.

Some prior literature has employed 2D representations for SITS pixels. For example, Wang and Oates~\citep{wang2015encoding} applied Gramian Summation Angular Fields (GASF) and Markov Transition Fields (MTF) to 12 time series datasets for classification tasks. Other examples include the studies that explored these representations in classification problems (e.g., regions with and without eucalyptus plantation)~\citep{dias2020multirepresentational, Dias2020GRSL, Menini2019GRSL}. Additionally, Abidi et al.~\citep{abidi2023combining} also investigated similar methodologies in their research, utilizing 2D representations to perform land cover classification in the French territory with a ResNet 50 network with pre-trained weights for feature extraction and a trained classifier.

Our contribution extends beyond the aforementioned works, primarily due to the multimodal and self-supervised nature of our application. Conducted experiments not only demonstrate the effectiveness of our methodology but also highlight the advantages of utilizing a richer, 2D representation of SITS data in conjunction with RSI. To the best of our knowledge, features extracted from those pixel representations have not been explored in self-supervision methodologies, a gap bridged in this paper.

\section{Pixel-wise Multimodal Contrastive Learning}
\label{sec:PIMC}

% Descrever que aqui falaremos como foi feita a implementar
This section relates to the first research question, focusing on how to perform multimodal contrastive learning using two-dimensional representations of time series and SITS. This section provides a detailed description of the proposed PIMC approach, including the construction of datasets with 2D representations of time series, the model architecture, and the multimodal contrastive training process.

\subsection{Overview}

In this paper, we investigate the use of 2D representations (e.g., recurrence plots) of time series derived from SITS vegetation indices in a multimodal contrastive self-supervised learning scheme. This method facilitates the creation of a unified feature space that accommodates both RSI and pixel-wise temporal data.
Since PIMC involves a multimodal training process, the two-dimensional time series representations serve as one modality while the RSI is the other. After the PIMC training, we have two distinct models, each tailored to its respective modality. 

\begin{figure*}[!t]
\centering
\includegraphics[width=1\linewidth]{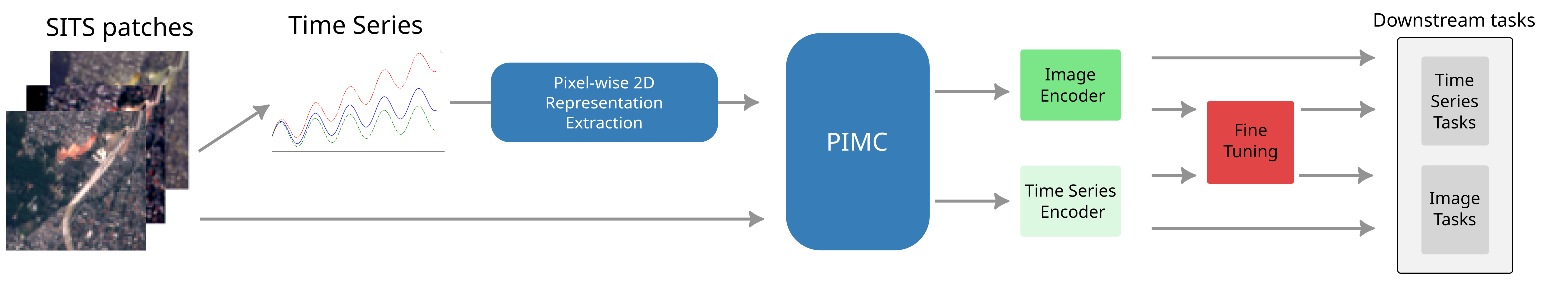}
\caption{The figure illustrates the pipeline of our multimodal self-supervision learning approach, wherein we start the process utilizing the patches from the SITS, where pixels are selected to construct the time series using the vegetation indices values. Subsequently, we extract the two-dimensional representation from the time series. Both modalities are employed in the training of the image encoder and time series encoder via the PIMC approach. The trained models and their fine-tuned versions are utilized in downstream tasks to assess the effectiveness of utilizing two-dimensional representations in training via PIMC.}
\label{fig:overview}
\end{figure*}

Figure~\ref{fig:overview} illustrates the entire process of self-supervision learning. The training process starts with the use of patches from the SITS and selecting specific pixels to construct a time series based on vegetation index values. Next, a two-dimensional representation from the time series is computed. Both data modalities are then used to train the image and time series encoders using the PIMC approach. The trained models, along with their fine-tuned versions, are subsequently applied to downstream tasks (time-series and image-based) to evaluate their effectiveness.

\subsection{Pixel-wise 2D Representation Extraction}

% Todo o caminho das imagens de MSI ate chegarem em RP.

\begin{figure*}[!t]
\centering
\includegraphics[width=1\linewidth]{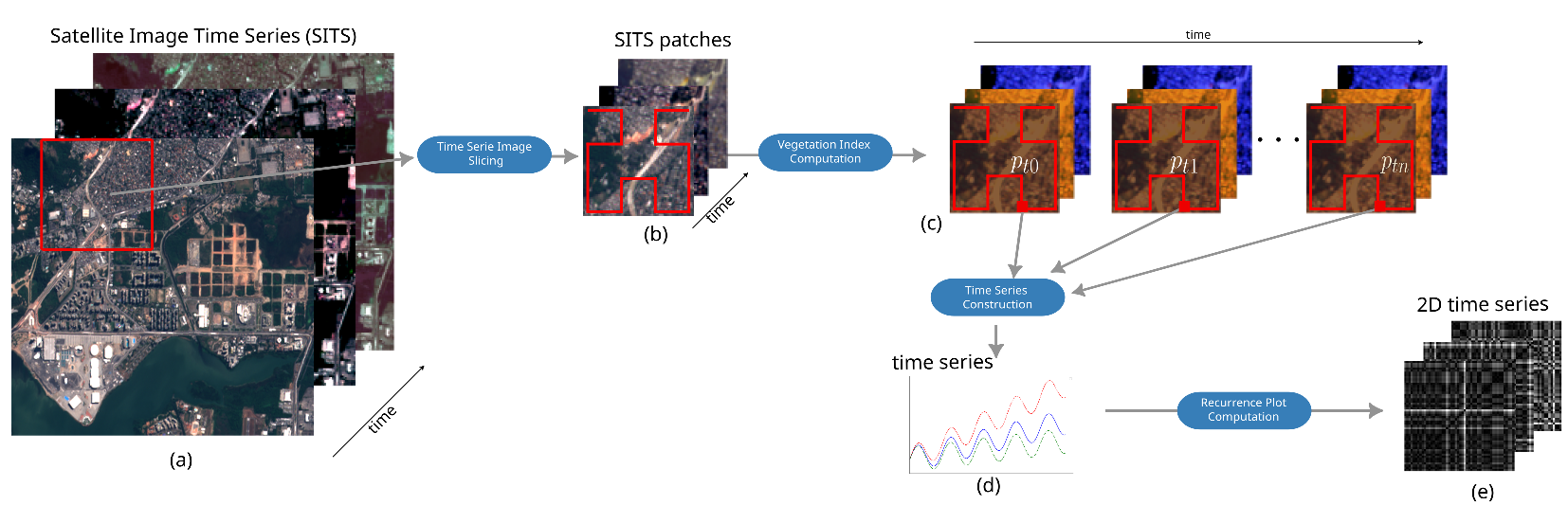}
\caption{Illustration of the creation of the 2D representation of pixel-wise vegetation indices from the SITS. (a) Process of dividing the SITS into patches; (b) process for computing the vegetation indices for each SITS patch; (c) selection of pixels within each image patch; (d) construction of 1D time series, representing the vegetation indices of each pixel at different timestamps; and (e) computation of recurrence plot, transforming the 1D time series into a 2D representation that encodes recurrent states in the time series.}
\label{fig:SITS_2_time_series}
\end{figure*}

Let $D$ be a dataset composed of $r$ regions, where each region is composed of $t$ images. We used these sequences of images to create the Time Series Images (TSI) for training the PIMC and benchmark models. Figure~\ref{fig:SITS_2_time_series} illustrates the entire process of creating the 2D representation from SITS. Initially, the RSI is divided into patches. For each patch, we compute the vegetation indices, and a Hilbert curve~\citep{jagadish1997analysis} is used to sample points that represent the region within the RSI patch. In the subsequent step, these selected pixels are converted into a time series using the vegetation indices of each SITS. The final step involves computing the recurrence plot of all 1D time series, resulting in our final 2D representation of the data.

A key aspect of the dataset construction is that two different sets of pixels are utilized for training the PIMC architecture and for training and evaluating the supervised methods. Specifically, while the PIMC training employs pixels sampled from Hilbert curves, random pixels are chosen within the patch for the supervised training and evaluation in the downstream tasks. This ensures that the PIMC training process is conducted using different data from what is used in the validation tasks.

\subsubsection{Vegetation Index Calculation} While creating a time series from each multispectral band in the image is the standard approach for processing MRSI and TSI data, some authors~\citep{dias2020multirepresentational} have demonstrated that using vegetation indices can refine the data and enhance feature extraction from the MSRSI. Following this strategy, we select three commonly used vegetation indices from the literature (NDVI~\citep{rouse1974monitoring}, EVI~\citep{huete2002overview}, and SAVI~\citep{huete1988soil}). For each time series, we transform the representation from $\mathbb{R}^{{p}\times{c}\times{t}}$ to $\mathbb{R}^{{p}\times{3}\times{t}}$, where $p$ is the number of patches, $c$ is the number of channels in the MSRSI, and $t$ is the number of timestamps. This new representation captures the temporal changes in vegetation indices.

An important reason for selecting these three vegetation indices relies on the validation using a downstream task involving different vegetation and crop types with diverse temporal profiles. Recall that our approach is generic and may consider the use of other indices, which may be more appropriate for specific downstream tasks. Moreover, most computer vision models are designed to work with RGB images, which typically have three channels. By aligning our input data with three-band images, we can leverage pre-trained models for various computer vision tasks.

\subsubsection{TSI Slicing} The dataset $D$ consists of $R_{1}, R_{2}, \ldots, R_{r}$ SITS, where $r$ is the number of regions associated with a sequence of images $t$ (time series of images). For each $R \in D$, we sliced the images into patches/windows of $ps\times ps$ pixels. Consequently, each TSI with dimensions $\mathbb{R}^{{t}\times{c}\times{h}\times{w}}$ is transformed into $\mathbb{R}^{{t}\times{p}\times{c}\times{ps}\times{ps}}$, where $t$ is the number of patches that compose the entire image.

\subsubsection{Time Series Construction} With the sliced TSI representation, we sample $n$ pixels from each image patch to create the time series. Each time series consists of the pixel values for the sequence of MSRSI present in $D$. As a result, samples are transformed from the $\mathbb{R}^{{t}\times{p}\times{c}\times{ps}\times{ps}}$ to a $\mathbb{R}^{{p}\times{c}\times{n}}$ space, where each time series of length of $n$ contains $c$ values corresponding to each MSRSI band. This transformation significantly reduces the data size while providing detailed information from the selected pixels.

\subsubsection{2D Representation Construction} The previous steps focused on processing time series (originally one-dimensional for each vegetation index) to obtain two-dimensional representations. This transformation allows for a new approach to enhance data refinement, as intrinsic patterns within the time series can be emphasized, making the representation richer in detail regarding the changes and correlations between the points (in this case, the pixels).

To make this conversion, we opted for the recurrence plot~\citep{eckmann1995recurrence}, a technique that generates a 2D representation from a 1D time series by using the paired distances between points and considering how often each value appears in the sequence. Equation~\ref{eq:recurrence_plot} shows how the recurrence plot is calculated.
\begin{equation}
     RP_{i,j} = \lVert \mathbf{x}_i - \mathbf{x}_j \rVert
     \label{eq:recurrence_plot}
\end{equation}
\noindent where $\mathbf{x}$ is the value of the vegetation index time series at timestamps $i$ and $j$ ($i, j \leq t$).

One recurrence plot was created for each calculated vegetation index. Thus, the time series $\mathbb{R}^{{p}\times{3}\times{n}}$, which represents all three vegetation indices, transforms into the shape $\mathbb{R}^{{p}\times{3}\times{n}\times{n}}$ after the recurrence plot computation.

\subsection{Contrastive Training}

Our multimodal contrastive training follows the approach introduced by CLIP~\citep{radford2021learning}, employing two encoders trained on data from different modalities. In our case, we selected two ResNet-18~\citep{he2016deep} models as encoders, as they are relatively simple convolutional neural network (CNN) architectures with fewer parameters compared to contemporary transformer-based models that have billions of parameters. Using a smaller model effectively demonstrates the potential of our idea, given that these models have limited capacity while also allowing for faster training and validation processes. This approach of utilizing a small network is inspired by state-of-the-art works in self-supervised learning (SSL) in remote sensing, such as SeCo~\citep{manas2021seasonal}, CACo~\citep{Mall_2023_CVPR_CACo}, and MoCo~\citep{he2020momentum}.

\begin{figure}[!t]
\centering
\includegraphics[width=1\linewidth]{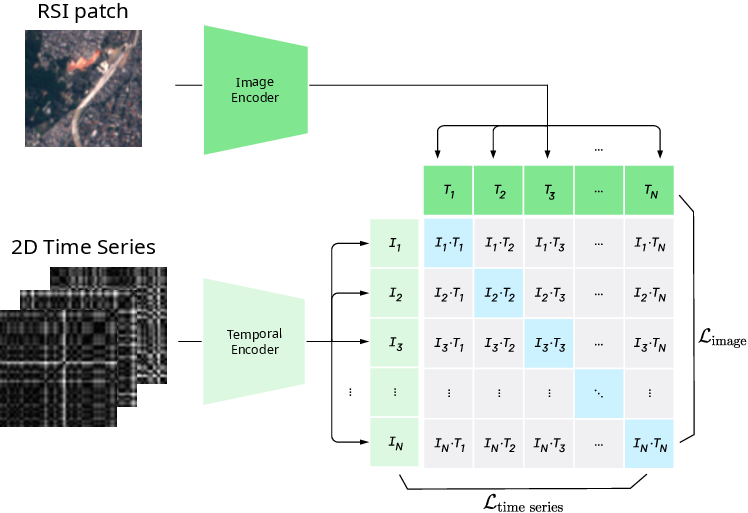}
\caption{Process of contrastive learning: (a) an RSI patch and (b) the combined 2D recurrence plots of the vegetation indices from one pixel within the RSI patch; and (c) the similarity matrix generated by the dot product between $\mathbf{I}$ and $\mathbf{T}$.}
\label{fig:CLIP_training}
\end{figure}

In our approach, the first encoder is responsible for extracting features from the RGB channels of the RSI images, denoted as $\mathbf{I}$, The second encoder processes the 2D recurrence plot representations derived from the time series of the pixels, with these features denoted as $\mathbf{T}$. Figure~\ref{fig:CLIP_training} illustrates how the training using both modalities works.

\subsubsection{Loss Calculation} The loss function is based on the original proposal by CLIP~\citep{radford2021learning}, where the similarity matrix $\mathbf{S}$ is calculated using the dot product between pairs of normalized feature spaces: $\mathbf{S} = \mathbf{I}_{\text{norm}} \times \mathbf{T}_{\text{norm}}^\top$. As our train aims to maximize similarity, we apply the cross-entropy loss~\citep{rumelhart1986learning} in both directions, using $\mathbf{S}$ and $\mathbf{S}^\top$. This ensures that both encoders compute effective features for the different modalities of data.

Thus, the two cross-entropy losses, $\mathcal{L}_{\text{image}}$ to estimate the alignment of the features from $\mathbf{I}$ and $\mathbf{T}$, and $\mathcal{L}_{\text{time series}}$ to estimate if the features from the two-dimensional representation $\mathbf{T}$ are aligned with the RSI $\mathbf{I}$, are calculated. This enables both models to learn how to represent the inputs in one modality similarly to the features from the other. The final loss $\mathcal{L}$ is computed as the mean of $\mathcal{L}_{\text{image}}$ and $\mathcal{L}_{\text{time series}}$, as expressed in Equation~\ref{eq:loss}.
\begin{equation}
\mathcal{L} = \frac{1}{2} \left(\mathcal{L}_{\text{image}} + \mathcal{L}_{\text{time series}}\right)
\label{eq:loss}
\end{equation}

\section{Experiments}
\label{sec:experiments}

This section relates to Research Questions 2 and 3. We conducted a set of experiments aiming to demonstrate the effectiveness of trained encoders using the PIMC methodology. In our experiments, we considered three downstream tasks involving both images and time series: land cover image classification, time-series-based vegetation index forecasting, and time-series-based pixel classification. In all tasks, we perform comparisons with well-established and SOTA methods.

The validation protocol was selected to evaluate the effectiveness of PIMC in two distinct ways. First, we assess the quality of encoders from both $\mathbf{I}$ and $\mathbf{T}$ on downstream tasks involving each of those modalities. Second, we evaluate the distribution of extracted features in the target feature space and compare them to conventional methods.

\begin{figure}[!t]
    \centering
    \includegraphics[width=0.8\linewidth]{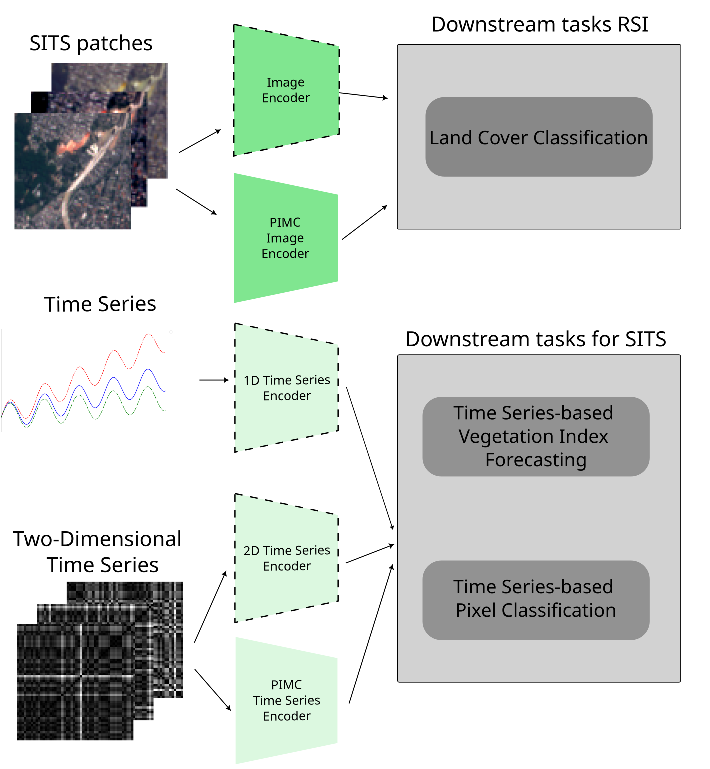}
    \caption{Illustration of the validation process for encoders generated using PIMC. The RSI images were processed as SITS patches and used as input for baseline encoders and the PIMC image encoder for the land cover classification downstream task. On the other hand, the time series data (1D and 2D representations) were used as input for the raw time series encoder and 2D time series encoder as (baselines),  and for the PIMC time series encoder. In this case, two downstream tasks are considered: time series-based vegetation index forecasting and time series-based pixel classification.}
    \label{fig:overview_experiments}
\end{figure}

Figure~\ref{fig:overview_experiments} illustrates the process of validating the encoders generated based on the proposed method. Encoders with dashed lines refer to baselines. For the land cover image classification downstream task, we compared the PIMC encoder with an image encoder trained in a supervised manner, the same network trained with the SSL state-of-the-art, and a vision transformer method. For the downstream task involving time series, we consider encoders created based on the raw time series (1D Time Series Encoder) and based on the created two-dimensional representations (2D Time Series Encoder), for which machine learning, convolutional network, and transformer-based methods were applied.

\subsection{PIMC Training Process}

\subsubsection{Dataset for Training}
The TSI is extracted from the Panoptic Agricultural Satellite TIme Series (PASTIS)~\citep{garnot2021panoptic} dataset, with 2,433 multispectral SITS with images of $128 \times 128$ pixels and $10$ bands. This dataset consists of images captured from fields in France, with 18 classes of plantations. This dataset was selected for training the PIMC because it contains multispectral bands, allowing us to calculate the vegetation indices, and also contains different types of vegetation and crops, thus ensuring greater diversity in the dataset.

\subsubsection{Training Details}
The PIMC comprises two encoders, the image encoder and the time series encoder. Both were implemented using the ResNet-18. To update the weights of both, the Adam~\citep{kingma2014adam} optimizer was employed with a learning rate of $1e^{-3}$ with a weight decay of $1e^{-4}$ for $400$ epochs on the PASTIS training set. The training set was selected according to the specifications of the study that introduced the PASTIS dataset~\citep{garnot2021panoptic} (folders 1, 2, and 3), while folders 4 and 5 were used for validation and testing, respectively. An NVIDIA Tesla A100 was utilized for the training process.

\subsection{Downstream Tasks: Protocol and Results}
\label{benchmarks}

As downstream tasks, we selected three different problems: time series-based pixel classification, time series-based vegetation index forecasting, and land cover classification.

\subsubsection{Time Series-Based Pixel Classification}

The pixel classification task was conducted using the PASTIS dataset, from which we extracted the ``class'' of each pixel based on the semantic segmentation masks representing the type of crop in those areas. To train the models, we selected 18 possible labels for the fields of the 20 originally proposed by the authors, excluding pixels categorized as background or uncertainty.

As baselines for the 1D time series encoder, we selected three different methods: Extreme Gradient Boosting (XGBoost)~\citep{chen2016xgboost} a machine learning algorithm based on gradient boosting technique, that uses decision trees to regression or classification tasks; 1D CNN composed of $5$ blocks of convolutional layers with ReLU activation and batch normalization; a Long Short-Term Memory (LSTM)~\citep{hochreiter1997lstm} model that is a recurrence neural network that works learning how to control the flow of information in sequential data, in our LSTM implementation contains $4$ with $64$ dimensions in its features and MOMENT~\citep{goswami2024moment} a foundation model trained in for general proposed in time series data.

The downstream task training was conducted for $100$ epochs, with the Adam optimizer. The results were obtained from the test set following the original PASTIS split, with $200$ pixels selected per SITS. The 1D time series encoder was trained in a supervised manner. MOMENT was applied as feature extraction and fine-tuned, while the two-dimensional time series representation encoder was just used as a feature extractor or fine-tuned with the dataset with labels. We evaluated the classification results using accuracy and balanced accuracy metrics. 

\begin{table*}[!t]
\centering
\caption{Time series pixel classification results considering encoders trained using 1D and two-dimensional time series representations, with and without fine-tuning (FT). The table shows the accuracy (ACC), balanced accuracy (BAL ACC), and F1 score metrics for a range of models.}
\label{tab:pixel_classification}
    \begin{tabular}{@{}lcccc@{}}
        \toprule
        \multicolumn{1}{c}{\textbf{Models}} & \textbf{Time Series Representation} & \multicolumn{1}{c}{\textbf{ACC} $\uparrow$} & \multicolumn{1}{c}{\textbf{BAL ACC} $\uparrow$} & \multicolumn{1}{c}{\textbf{F1} $\uparrow$} \\ 
        \midrule
        XGBoost                            & 1D & 55.80 & 23.67 & 0.35 \\
        1D CNN                             & 1D & 59.42 & 21.81 & 0.38 \\
        LSTM                               & 1D & 59.92 & 40.73 & 0.41 \\
        MOMENT~\citep{goswami2024moment}    & 1D & 59.92 & 40.73 & 0.41 \\
        MOMENT$^{\text{FT}}$~\citep{goswami2024moment} & 1D & 61.12 & 42.73 & 0.43 \\
        \midrule
        ImageNet                           & 2D & 21.78 & 20.76 & 0.19 \\
        SeCo~\citep{manas2021seasonal}      & 2D & 25.15 & 18.31 & 0.21 \\
        ViT 32~\citep{manas2021seasonal}    & 2D & 25.15 & 18.31 & 0.21 \\
        DINO MC~\citep{goswami2024moment}   & 2D & 35.33 & 26.19 & 0.29 \\
        \textbf{PIMC}                      & 2D & 25.95 & 20.42 & 0.22 \\ 
        ImageNet$^{\text{FT}}$                      & 2D & 63.72 & 28.51 & 0.45 \\
        SeCo$^{\text{FT}}$~\citep{manas2021seasonal} & 2D & 70.15 & 45.16 & 0.48 \\
        ViT 32$^{\text{FT}}$~\citep{manas2021seasonal} & 2D & 69.32 & 43.11 & 0.48 \\
        DINO MC$^{\text{FT}}$~\citep{goswami2024moment} & 2D & 64.38 & \textbf{60.10} & 0.49 \\
        \textbf{PIMC$^{\text{FT}}$}                 & 2D & \textbf{71.07} & 45.96 & \textbf{0.50} \\ 
        \bottomrule
    \end{tabular}
\end{table*}

Table~\ref{tab:pixel_classification} shows the classification results for the 1D and 2D methods. As we can observe, models trained using two-dimensional representations (e.g., ImageNet, SeCo, ViT 32, DINO MC, and PIMC) performed better when compared with those trained on raw time series after the fine-tuning (e.g., ImageNet$^{FT}$, SeCo$^{FT}$, ViT 32$^{FT}$, DINO MC$^{FT}$, and PIMC$^{FT}$). Also, we can observe that the use of fine-tuning strategies led to improved results. Finally, the PIMC-based encodings led to the highest ACC and F1 classification scores (in bold).

\begin{figure}[!t]
    \centering
    \subfloat[PIMC$^{\text{FT}}$]{%
        \includegraphics[width=0.32\textwidth]{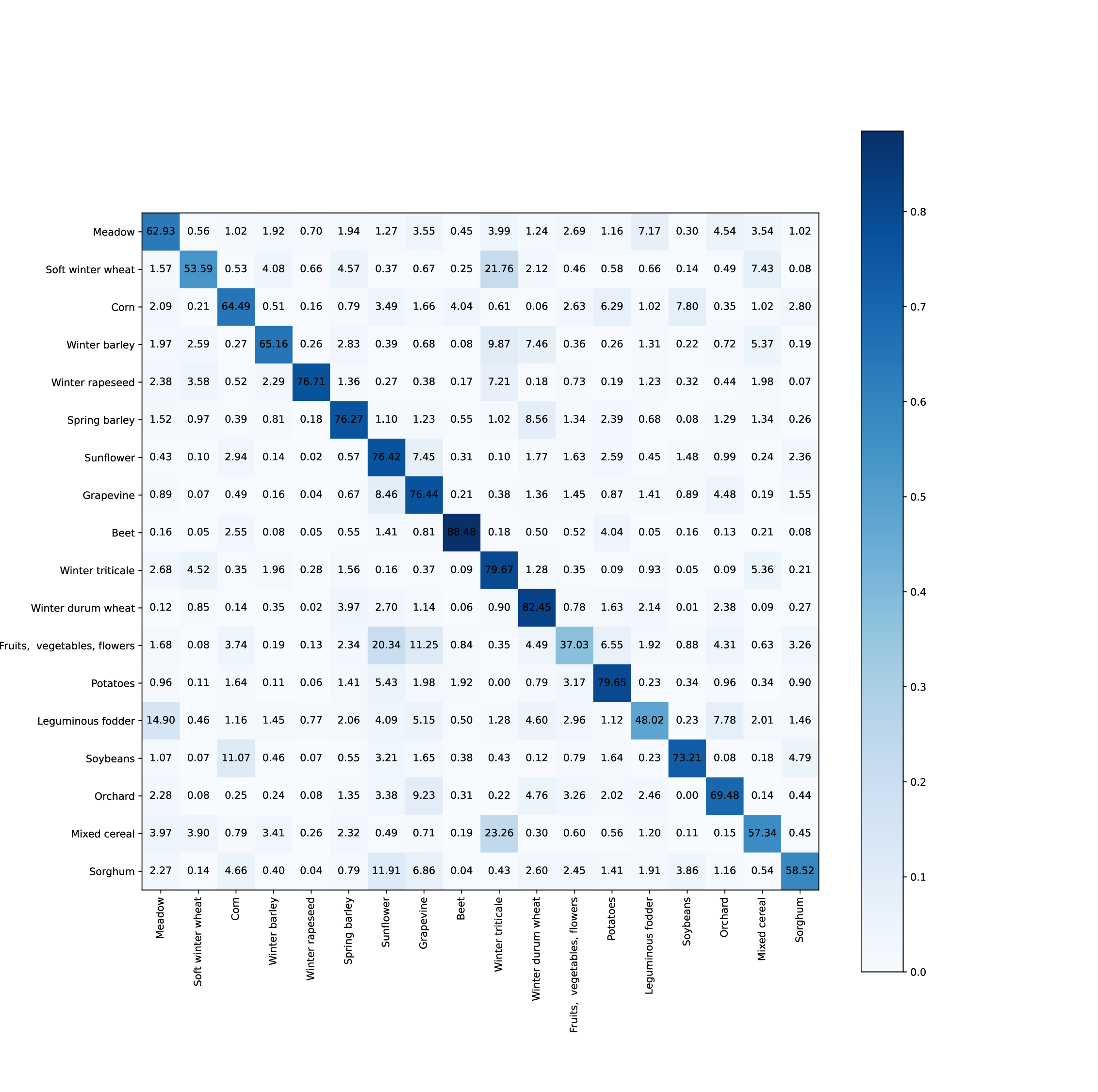}
    }\hfill
    \subfloat[DINO$^{\text{FT}}$]{%
        \includegraphics[width=0.32\textwidth]{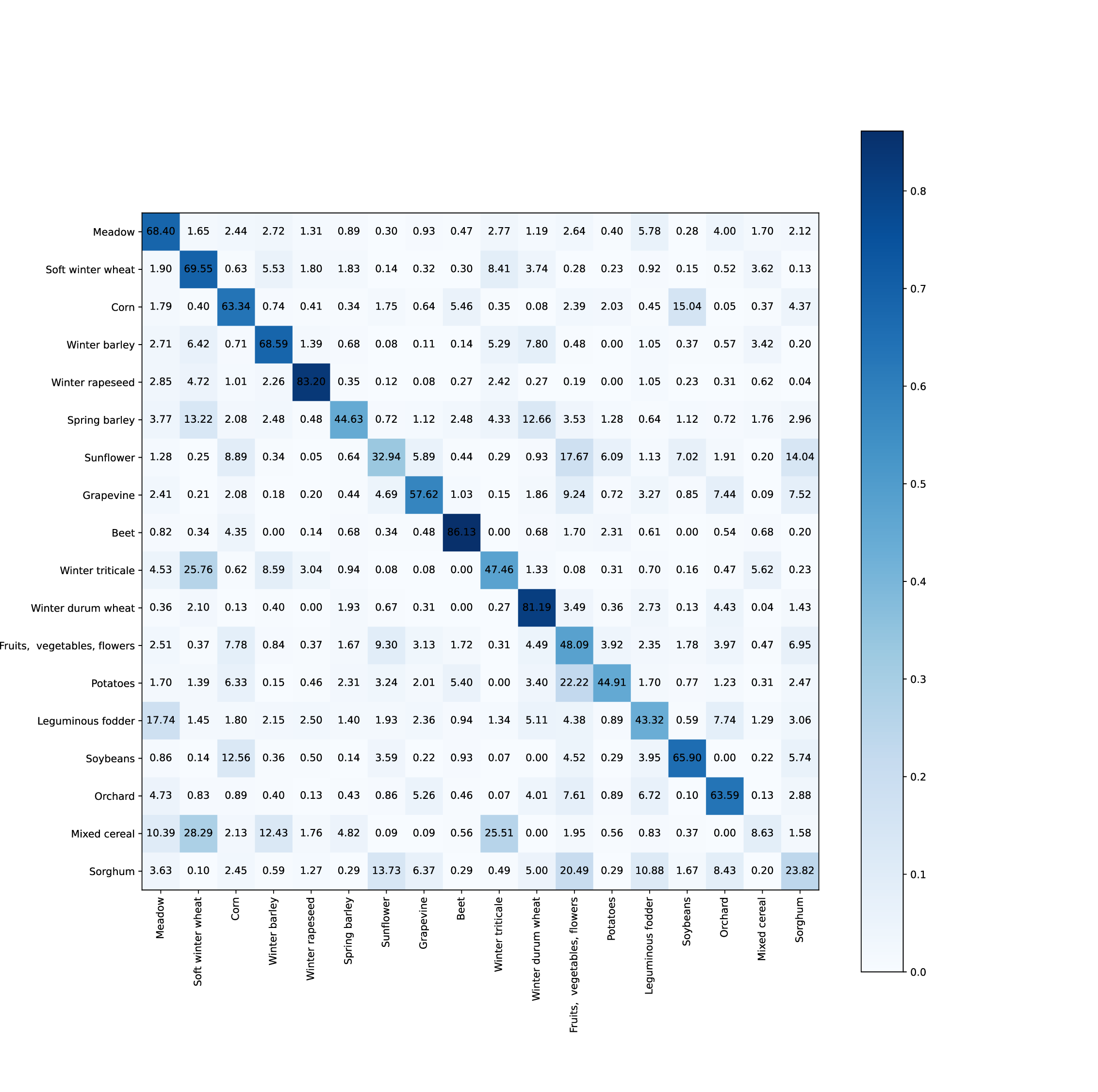}
    }\hfill
    \subfloat[SeCo$^{\text{FT}}$]{%
        \includegraphics[width=0.32\textwidth]{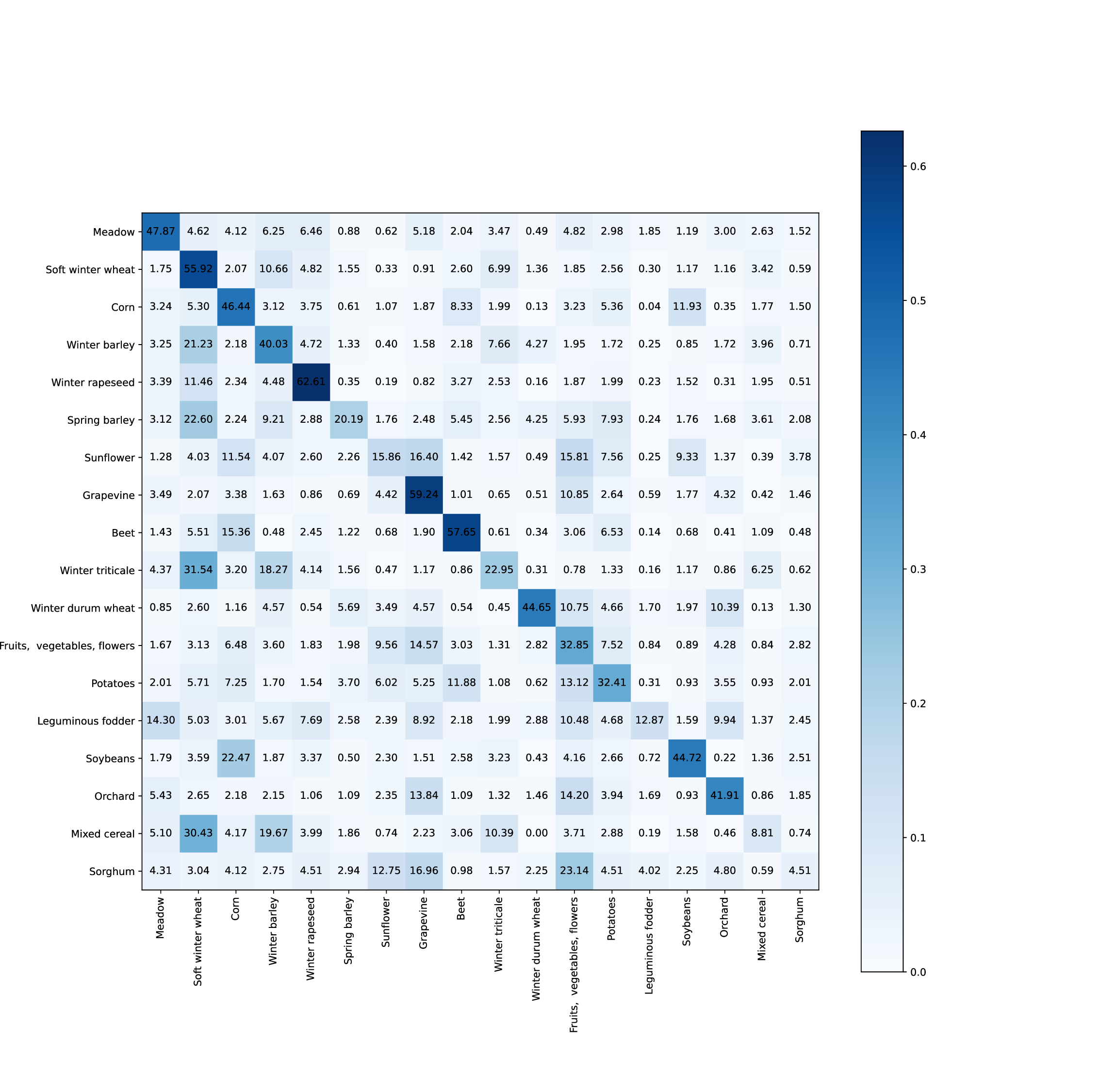}
    }
    \caption{Confusion matrices for the PIMC$^{\text{FT}}$, DINO$^{\text{FT}}$ and SeCo$^{\text{FT}}$ classification models. (a) PIMC fine-tuned; (b) DINO MC fine-tuned; (c) SeCo fine-tuned. The PIMC model shows reduced confusion between classes.}
    \label{fig:pixel_classification}
\end{figure}

The dataset exhibits a highly imbalanced class distribution, making it more challenging for the model to accurately distinguish between classes. For example, class Meadow has more than $100,000$ samples for training, while class Leguminous fodder has only $6,000$. This behavior is illustrated in Figure~\ref{fig:pixel_classification}, where the confusion matrices of the best-performing models (PIMC$^{\text{FT}}$ and SeCo$^{\text{FT}}$ reflect the difficulties in achieving perfect separation among the classes.

The models fine-tuned with labels using the PIMC method demonstrated superior separation capabilities. Additionally, both models outperformed all counterparts trained with 1D representations, indicating the enhanced quality of the computer vision models employed. This evidence supports the assertion that 2D representations provide more informative features for the models. Furthermore, the results obtained with PIMC showcased higher accuracy than one of the state-of-the-art self-supervised learning (SSL) models in remote sensing applications.

However, the results obtained using only the PIMC, SeCo, ViT, DINO MC, and ImageNet weights without fine-tuning indicate that extracting features in a zero-shot scenario from 2D representations poses a significant challenge for the models. After fine-tuning, the 2D models surpassed the 1D models trained in a supervised manner.

\subsubsection{Time Series-Based Vegetation Index Forecasting} 
This task aims to predict 10 subsequent values of a vegetation index time series, based on an input time series of size 32.

As baselines, we again use the same models used in the previous task (1D CNN, LSTM, MOMENT, and XGBoost), now trained to process the vegetation indices in their original 1D (raw) format. The goal is to assess how the use of a 2D representation can enhance the predictive capacity of encoders.

The training protocol employed is the same as for the time series-based pixel classification, with $100$ epochs of training the 1D time series encoders and fine-tuning of the models with pre-trained weights. To evaluate the quality of the forecasting, the mean absolute error (MAE) and mean squared error (MSE) were employed.

\begin{table*}[!t]
\centering
\caption{Time series forecasting results considering encoders trained using 1D and 2D time series representations, with and without fine-tuning (FT). Mean absolute error (MAE), mean square error (MSE), and root mean square error (RMSE) for NDVI, EVI, and SAVI.}
\label{tb:forecast_results}

\small
\setlength{\tabcolsep}{3pt}
\renewcommand{\arraystretch}{1.1}

\begin{adjustbox}{max width=\textwidth}
\begin{tabular}{lcccccccccc}
\toprule
\textbf{Model} & \textbf{Rep.} &
\multicolumn{3}{c}{\textbf{NDVI}} &
\multicolumn{3}{c}{\textbf{EVI}} &
\multicolumn{3}{c}{\textbf{SAVI}} \\

\cmidrule(lr){3-5} \cmidrule(lr){6-8} \cmidrule(lr){9-11}
 & & RMSE$\downarrow$ & MSE$\downarrow$ & MAE$\downarrow$
   & RMSE$\downarrow$ & MSE$\downarrow$ & MAE$\downarrow$
   & RMSE$\downarrow$ & MSE$\downarrow$ & MAE$\downarrow$ \\
\midrule
XGB & 1D & 0.6789 & 0.4609 & 0.5372 & 0.7136 & 0.5093 & 0.5725 & 0.7231 & 0.5229 & 0.5684\\
1D CNN & 1D & 0.6808 & 0.4635 & 0.6245 & 0.6392 & 0.4085 & 0.5705 & 0.6467 & 0.4182 & 0.5897\\
LSTM & 1D & 0.8603 & 0.7401 & 0.7011 & 0.8258 & 0.6820 & 0.6784 & 0.7999 & 0.6399 & 0.6327\\
MOMENT & 1D & 0.5829 & 0.3398 & 0.5063 & 0.5900 & 0.3481 & 0.5052 & 0.6052 & 0.3663 & 0.5205\\
MOMENT$^{\text{FT}}$ & 1D & 0.5743 & 0.3298 & 0.4794 & 0.5951 & 0.3541 & 0.4951 & 0.6187 & 0.3828 & 0.5150\\
\midrule
ImageNet & 2D & 0.6919 & 0.4788 & 0.6321 & 0.6510 & 0.4239 & 0.5825 & 0.6801 & 0.4626 & 0.6086\\
SeCo & 2D & 0.6844 & 0.4684 & 0.6277 & 0.6452 & 0.4163 & 0.5760 & 0.6377 & 0.4067 & 0.5785\\
ViT-32 & 2D & 0.5885 & 0.3463 & 0.4896 & 0.5615 & 0.3153 & 0.4698 & 0.5472 & 0.2994 & 0.4535\\
DINO MC & 2D & 0.6053 & 0.3664 & 0.5130 & 0.5762 & 0.3320 & 0.4859 & 0.5574 & 0.3107 & 0.4685\\
\textbf{PIMC} & 2D & 0.6844 & 0.4684 & 0.6277 & 0.6452 & 0.4163 & 0.5761 & 0.6377 & 0.4067 & 0.5785\\
\midrule
ImageNet$^{\text{FT}}$ & 2D & 0.7089 & 0.5025 & 0.6304 & 0.6530 & 0.4264 & 0.5811 & 0.6411 & 0.4110 & 0.5832\\
SeCo$^{\text{FT}}$ & 2D & 0.8552 & 0.7314 & 0.7027 & 0.7439 & 0.5533 & 0.6250 & 0.7788 & 0.6065 & 0.6487\\
ViT-32$^{\text{FT}}$ & 2D & 0.5885 & 0.3463 & 0.4896 & 0.5615 & 0.3153 & 0.4698 & 0.5472 & 0.2994 & 0.4535\\
DINO MC$^{\text{FT}}$ & 2D & 0.5147 & 0.2650 & 0.4007 & \textbf{0.5067} & 0.2568 & 0.4105 & \textbf{0.4907} & 0.2407 & 0.3937\\
\textbf{PIMC$^{\text{FT}}$} & 2D & \textbf{0.4849} & \textbf{0.2522} & \textbf{0.3947} & 0.5464 & \textbf{0.2481} & \textbf{0.3981} & 0.5387 & \textbf{0.2295} & \textbf{0.3789} \\
\bottomrule
\end{tabular}
\end{adjustbox}
\end{table*}

Table~\ref{tb:forecast_results} presents the prediction results for encoders using both the 1D and 2D representations, considering $10$ predicted values. Considering the prediction results, we can observe that the PIMC-based encoders process outperforms the ImageNet weights and achieves comparable results to those from SeCo and MOMENT. Additionally, all models based on two-dimensional representations led to superior predictive performance across all metrics.

\begin{figure}[!t]
\centering
\includegraphics[width=1\linewidth]{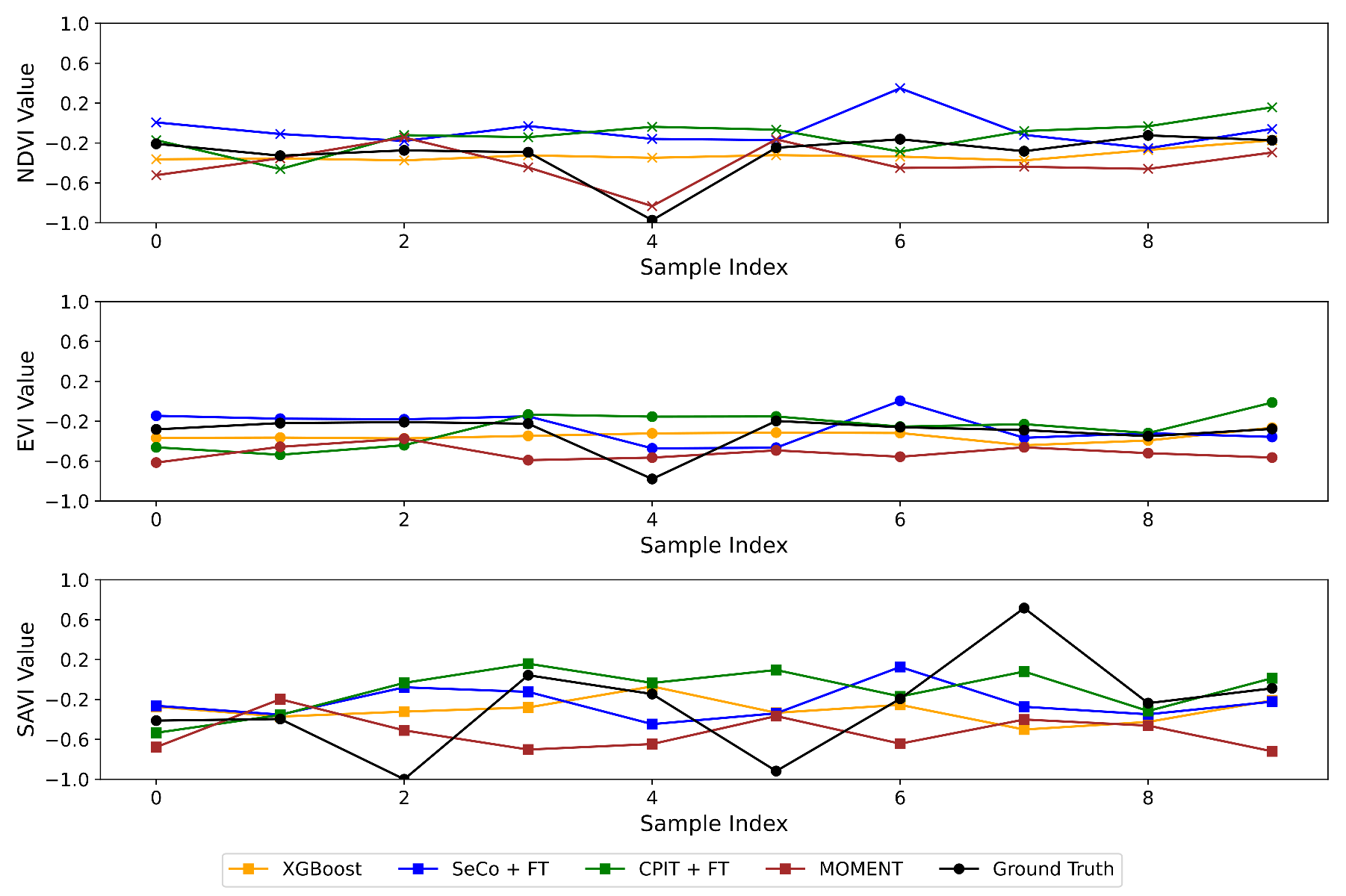}
\caption{Predictions for a sequence of $10$ timestamps for the three vegetation indices, evaluated across three different models. The first row displays the results for NDVI, the second row for EVI, and the third row for SAVI.}
\label{fig:pixel_forecast}
\end{figure}

The consistency of the predictions in the time series is illustrated in Figure~\ref{fig:pixel_forecast}, where the best model for each model predicts a sequence of $10$ samples for the three vegetation indices. From this, we can observe that the PIMC model led to closer results when compared to the ground truth of the time series.

It can be observed that the PIMC model more accurately follows the ground truth line for the predictions across all vegetation indices. Moreover, the performance of the encoder based on the 2D representation still outperforms that of the 1D representation, which not only deviates significantly from the ground truth but also displays greater instability in the predictions when compared with the ground truth.

\subsubsection{Land-Cover Classification} 
The classification of land cover is a frequently utilized method for evaluating the effectiveness of feature extraction methods based on self-supervised learning (SSL) models for remote sensing. To achieve accurate results, it is essential to have robust feature representations. The process entails the analysis of a remote sensing image (RSI) through the utilization of an encoder to extract meaningful features. The extracted features are then fed into a multi-layer perceptron (MLP), which is trained to classify the type of land cover or soil present in the image. This may include forests, water bodies, urban areas, or agricultural fields.

A prominent dataset for this purpose is EuroSAT~\citep{helber2019eurosat}, which consists of 27,000 labeled images spanning 10 classes. We used the same training procedures employed in previous experiments. In this specific task, however, we assess the performance of the PIMC-based image encoder. To assess model quality, we again relied on accuracy and balanced accuracy metrics.

Table~\ref{tab:eurosat_results} presents the land cover classification results for the evaluated models. As we can observe, 
the encoder we trained using PIMC showed strong performance in classification tasks, demonstrating PIMC's ability to create a shared feature space for both data types. This effectiveness was evident when the encoder was used for transfer learning and fine-tuned on this specific dataset.

\begin{table*}[!t]
    \centering
    \caption{Land-cover classification results considering encoders trained using two-dimensional time series representations, with and without fine-tuning (FT). The table shows the accuracy (ACC) and balanced accuracy (BAL ACC) metrics for a range of models.}
    \label{tab:eurosat_results}
    \begin{tabular}{@{}lccc@{}}
        \toprule
        \multicolumn{1}{c}{\textbf{Model}} & \textbf{Time Series representation} & \multicolumn{1}{c}{\textbf{ACC} $\uparrow$} & \multicolumn{1}{c}{\textbf{BAL ACC} $\uparrow$} \\ \midrule
        ImageNet                                & 2D & 78.38 & 77.93 \\
        ViT 32                                  & 2D & 82.24 & 81.72 \\
        SeCo~\citep{manas2021seasonal}          & 2D & 82.24 & 81.72 \\
        DINO MC~\citep{wanyan2023dinomc}         & 2D & 81.29 & 81.19 \\
        \textbf{PIMC}                           & 2D & 82.44 & 82.21 \\ 
        ImageNet$^{\text{FT}}$                           & 2D & 92.50 & 92.31 \\
        ViT 32$^{\text{FT}}$                             & 2D & 82.24 & 81.72 \\
        SeCo$^{\text{FT}}$~\citep{manas2021seasonal}      & 2D & 93.33 & 92.09 \\
        DINO MC$^{\text{FT}}$~\citep{wanyan2023dinomc}    & 2D & 96.10 & 95.41 \\
        \textbf{PIMC$^{\text{FT}}$}                      & 2D &\textbf{97.44} &\textbf{97.36} \\ \bottomrule
    \end{tabular}
\end{table*}

\subsection{Qualitative Assessment}

The experiments comparing the 1D and 2D representations of the time series reveal a significant positive impact on the training process when utilizing the 2D representation. The results from supervised training demonstrate superior performance compared to 1D supervised training with similar architectures. Additionally, the 2D representation showed itself to be a more effective pre-trained model for fine-tuning on both 1D and 2D data.

\subsubsection{Cases of Failure and Success}

\begin{figure}[!ht]
\centering
\subfloat[Meadow vs. Leguminous]{\includegraphics[width=0.8\columnwidth]{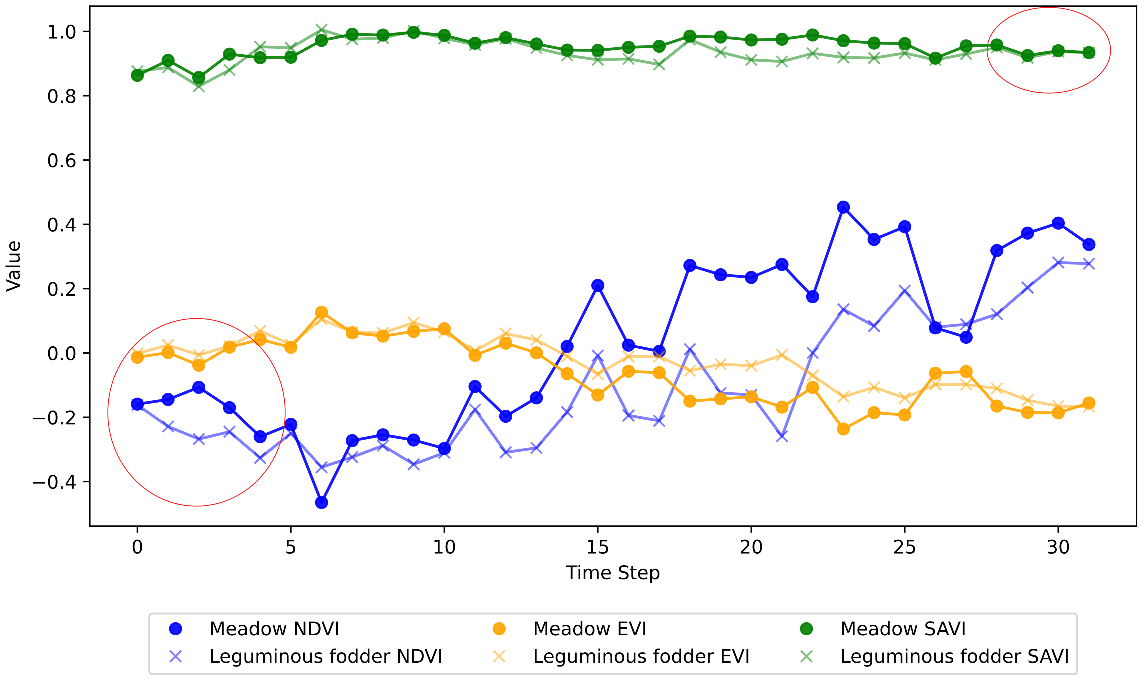}} \\ 
\subfloat[Meadow vs. Corn]{\includegraphics[width=0.8\columnwidth]{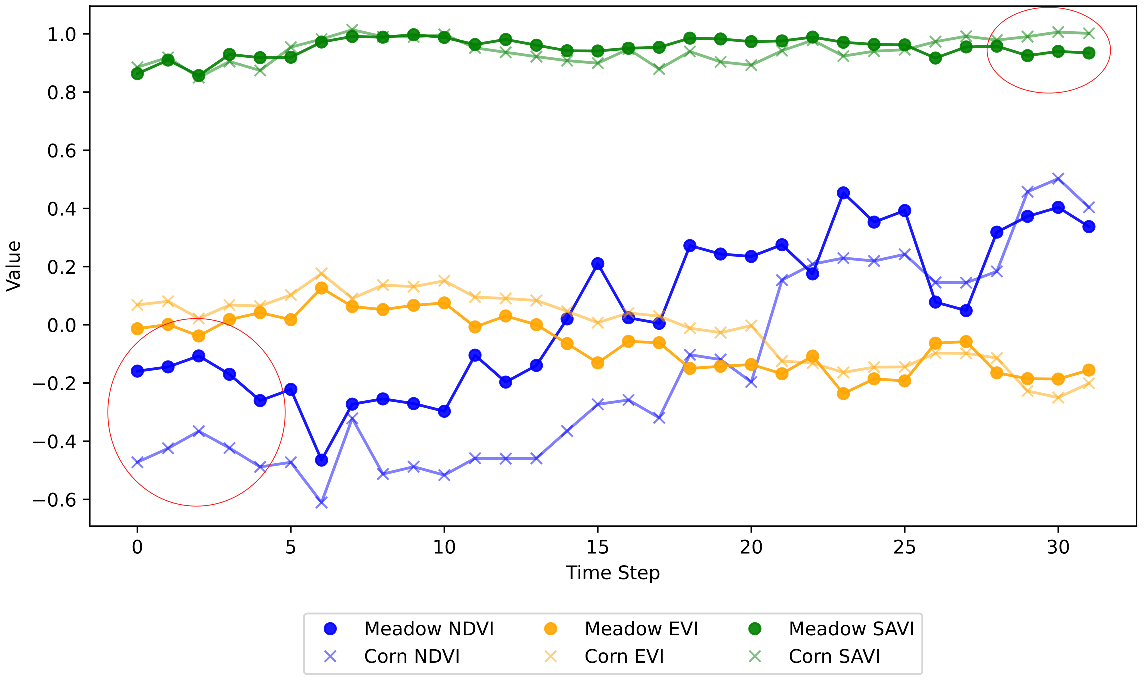}} 
\caption{Temporal behavior of the vegetation indices for the classes Meadow vs. Leguminous fodder and Meadow vs. Corn. The red-highlighted points in the time series illustrate our observation that the similarity in behaviors can complicate accurate predictions.}
\label{fig:indices_similarity}
\end{figure}

Despite the overall success of our approach, certain failure cases were identified. By analyzing the confusion matrix shown in Figure~\ref{fig:pixel_classification}, we determined which classes caused the most confusion during the model's classification process. The classes exhibited very similar patterns in their vegetation indices, as illustrated in Figure~\ref{fig:indices_similarity}.

% \begin{figure}[!htb]
% \centering
% \includegraphics[width=1\linewidth]{images/indicies_similarity.eps}
% \caption{Temporal behavior of the vegetation indices for the classes Meadow vs. Leguminous fodder and Meadow vs. Corn. The red-highlighted points in the time series illustrate our observation that the similarity in behaviors can complicate accurate predictions.}
% \label{fig:indices_similarity}
% \end{figure}

Since Meadow is the most prevalent class in the dataset, the classes that exhibit significant similarity in the time series tend to have comparable features. Consequently, the distribution of samples often leads the model to classify instances as the predominant class.

The PIMC method has shown potential in training models to effectively represent data from 2D representations of time series and the original RSI, as evidenced by its superior classification performance for classes with fewer labels compared to other approaches.

Furthermore, the 2D representation can be improved to address certain failure cases, particularly in the classification of challenging classes across all downstream tasks. For instance, incorporating additional vegetation indices could enhance information representation, as some soil types and crops may exhibit similar coloration when only using NDVI, EVI, and SAVI.

\begin{figure}[!ht]
    \centering
    % Row 1 - Permanent Crop
    \subfloat[Perman. Crop]{%
        \begin{tikzpicture}
            \node[draw=red, line width=1pt, rounded corners] {
                \includegraphics[width=0.2\columnwidth]{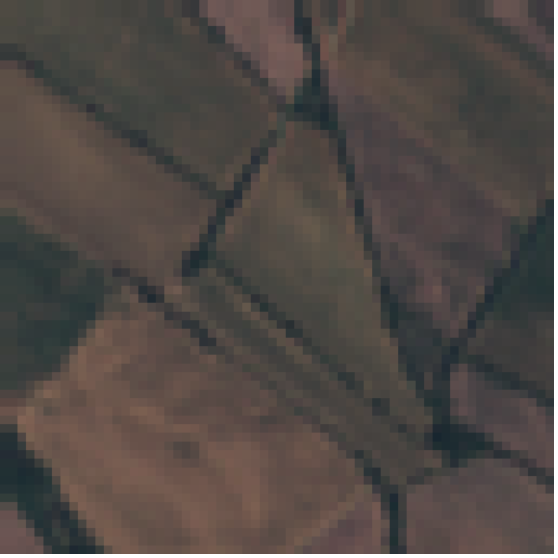}
            };
        \end{tikzpicture}%
        \label{fig:similar_permanentCrop_0}
    }
    \hfill
    \subfloat[Perman. Crop]{%
        \begin{tikzpicture}
            \node[draw=red, line width=1pt, rounded corners] {
                \includegraphics[width=0.2\columnwidth]{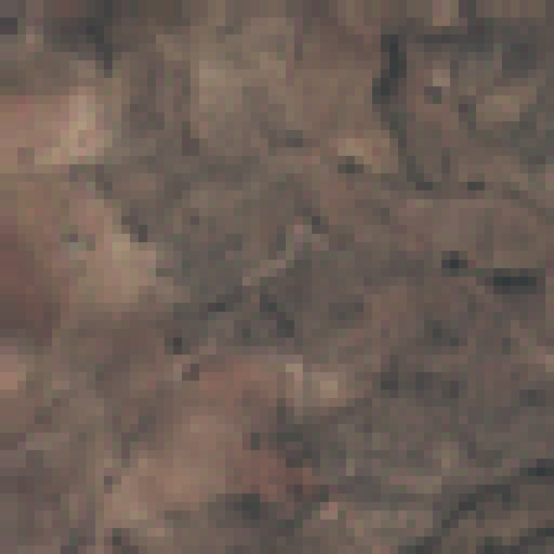}
            };
        \end{tikzpicture}%
        \label{fig:similar_permanentCrop_2}
    }
    \hfill
    \subfloat[Perman. Crop]{%
        \begin{tikzpicture}
            \node[draw=red, line width=1pt, rounded corners] {
                \includegraphics[width=0.2\columnwidth]{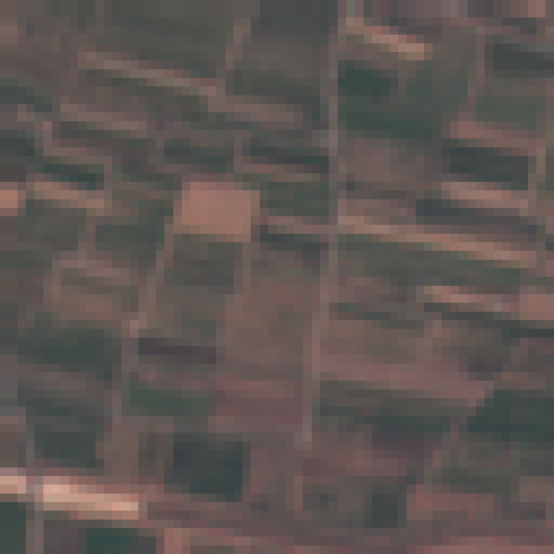}
            };
        \end{tikzpicture}%
        \label{fig:similar_permanentCrop_3}
    }
    \hfill
    \subfloat[Perman. Crop]{%
        \begin{tikzpicture}
            \node[draw=red, line width=1pt, rounded corners] {
                \includegraphics[width=0.2\columnwidth]{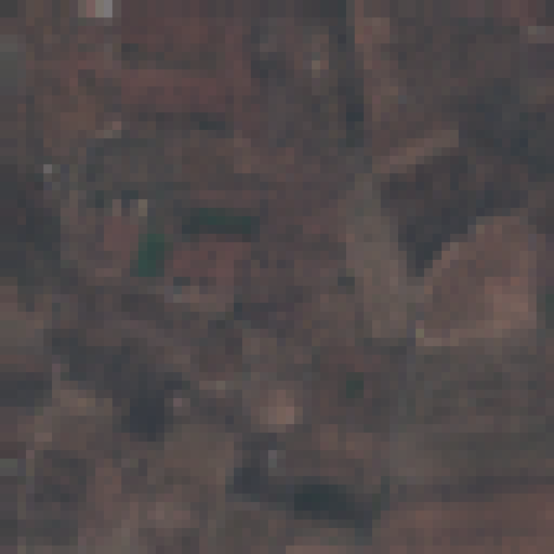}
            };
        \end{tikzpicture}%
        \label{fig:similar_permanentCrop_4}
    } \\

    % Row 2 - Pasture
    \subfloat[Pasture]{%
        \begin{tikzpicture}
            \node[draw=red, line width=1pt, rounded corners] {
                \includegraphics[width=0.2\columnwidth]{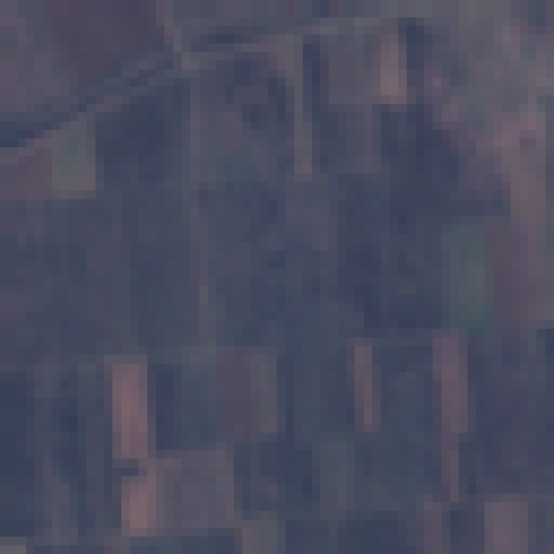}
            };
        \end{tikzpicture}%
        \label{fig:similar_pasture_0}
    }
    \hfill
    \subfloat[Pasture]{%
        \begin{tikzpicture}
            \node[draw=red, line width=1pt, rounded corners] {
                \includegraphics[width=0.2\columnwidth]{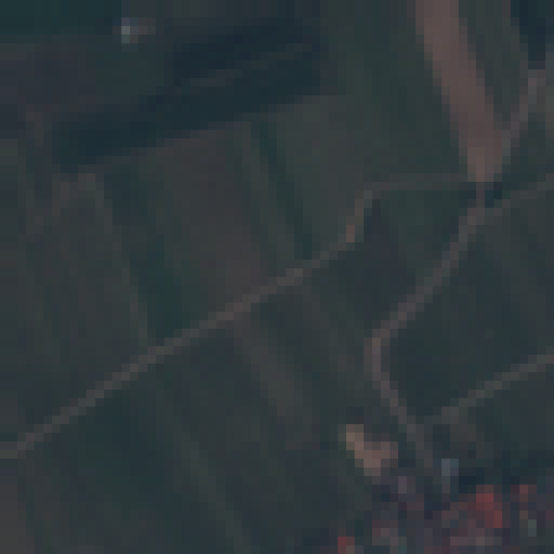}
            };
        \end{tikzpicture}%
        \label{fig:similar_pasture_1}
    }
    \hfill
    \subfloat[Pasture]{%
        \begin{tikzpicture}
            \node[draw=red, line width=1pt, rounded corners] {
                \includegraphics[width=0.2\columnwidth]{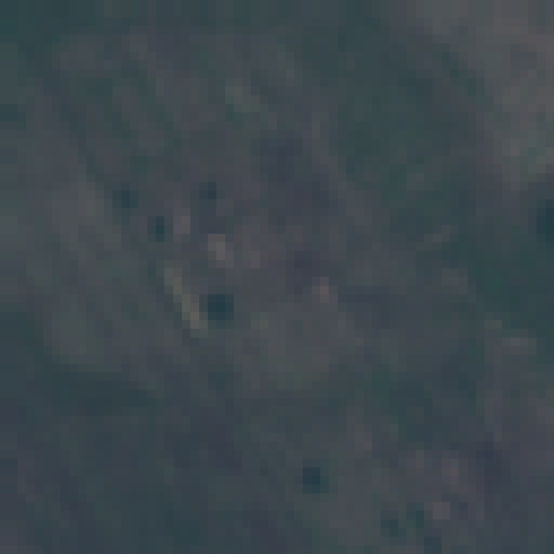}
            };
        \end{tikzpicture}%
        \label{fig:similar_pasture_2}
    }
    \hfill
    \subfloat[Pasture]{%
        \begin{tikzpicture}
            \node[draw=red, line width=1pt, rounded corners] {
                \includegraphics[width=0.2\columnwidth]{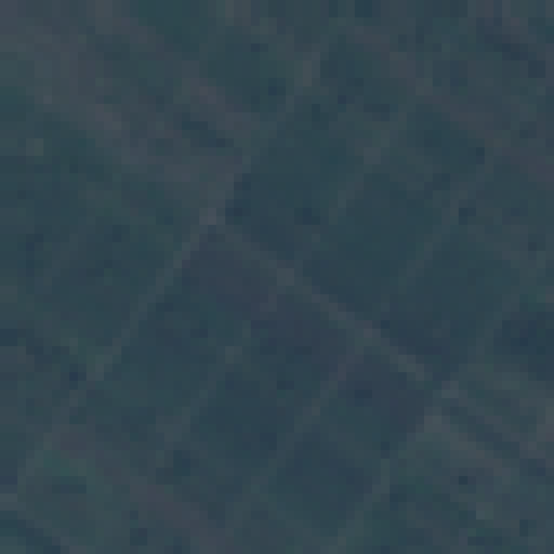}
            };
        \end{tikzpicture}%
        \label{fig:similar_pasture_3}
    } \\

    % Row 3 - Herbaceous Vegetation
    \subfloat[Herb. Veg.]{%
    \begin{tikzpicture}
            \node[draw=green, line width=1pt, rounded corners] {
                \includegraphics[width=0.20\columnwidth]{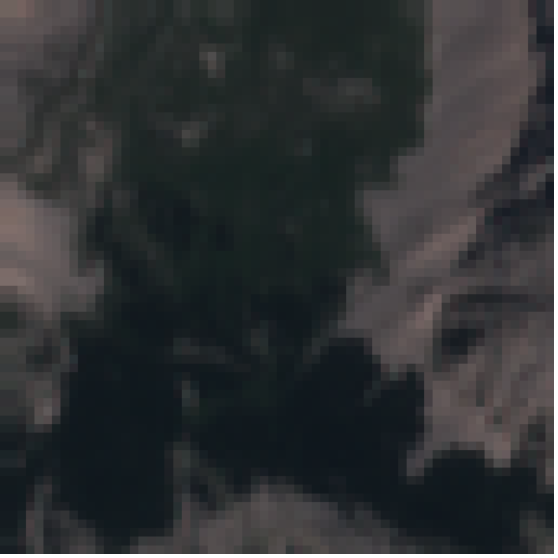}
            };
        \end{tikzpicture}%
        \label{fig:similar_herbacious_0}
    }
    \hfill
    \subfloat[Herb. Veg.]{%
        \begin{tikzpicture}
            \node[draw=green, line width=1pt, rounded corners] {
                \includegraphics[width=0.20\columnwidth]{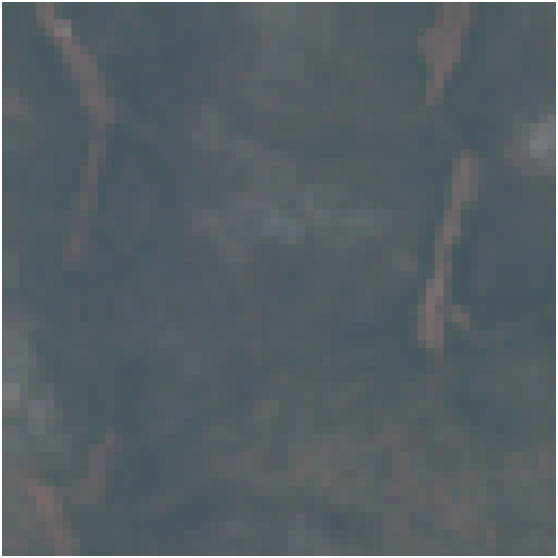}
            };
        \end{tikzpicture}%
        \label{fig:similar_herbacious_1}
    }
    \hfill
    \subfloat[Herb. Veg.]{%
        \begin{tikzpicture}
            \node[draw=green, line width=1pt, rounded corners] {
                \includegraphics[width=0.20\columnwidth]{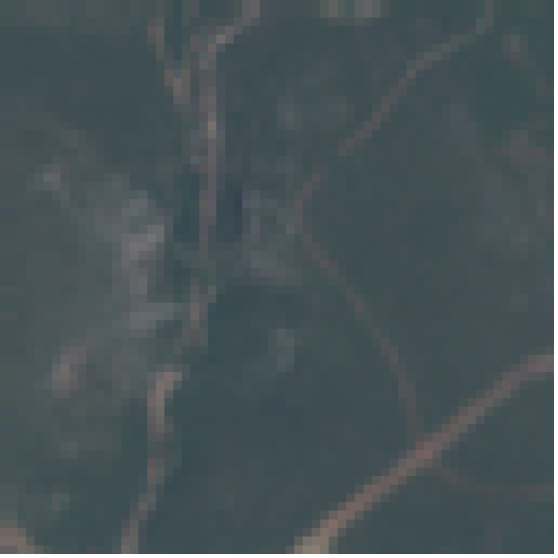}
            };
        \end{tikzpicture}%
        \label{fig:similar_herbacious_2}
    }
    \hfill
    \subfloat[Herb. Veg.]{%
        \begin{tikzpicture}
            \node[draw=green, line width=1pt, rounded corners] {
                \includegraphics[width=0.20\columnwidth]{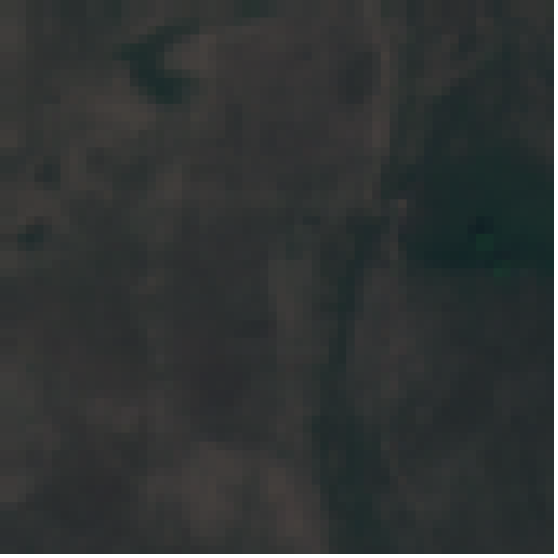}
            };
        \end{tikzpicture}%
        \label{fig:similar_herbacious_3}
    } \\
    
    % Row 4 - Residential
    \subfloat[Residential]{%
        \begin{tikzpicture}
            \node[draw=green, line width=1pt, rounded corners] {
                \includegraphics[width=0.20\columnwidth]{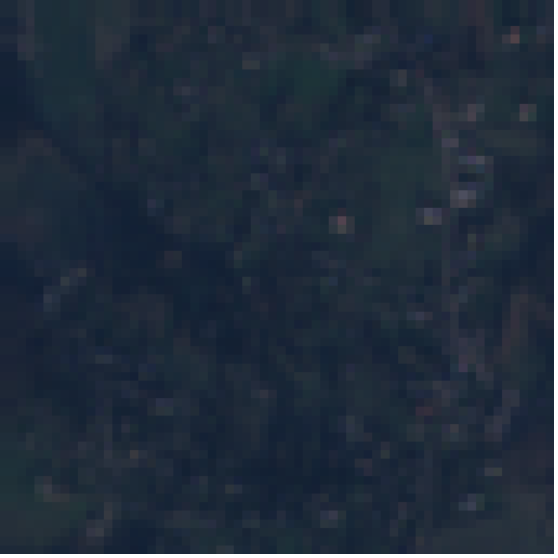}
            };
        \end{tikzpicture}%
        \label{fig:similar_residential_0}
    }
    \hfill
    \subfloat[Residential]{%
        \begin{tikzpicture}
            \node[draw=green, line width=1pt, rounded corners] {
                \includegraphics[width=0.20\columnwidth]{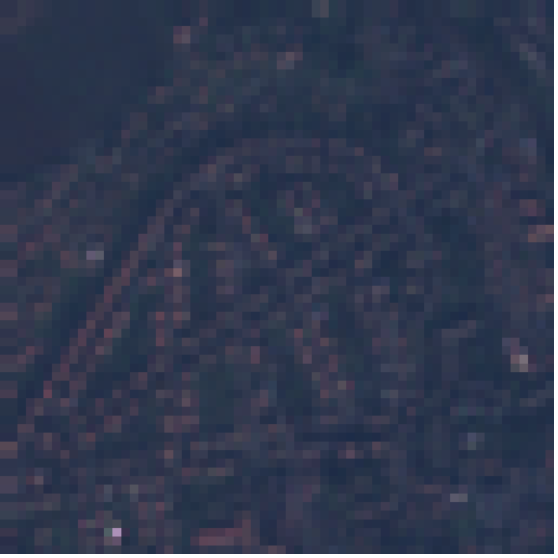}
            };
        \end{tikzpicture}%
        \label{fig:similar_residential_1}
    }
    \hfill
    \subfloat[Residential]{%
        \begin{tikzpicture}
            \node[draw=green, line width=1pt, rounded corners] {
                \includegraphics[width=0.20\columnwidth]{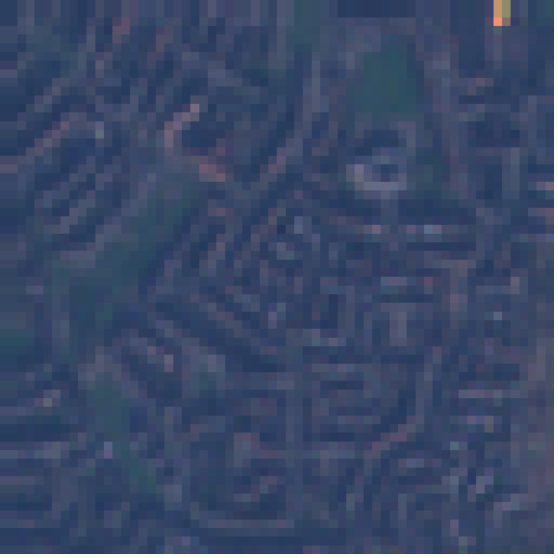}
            };
        \end{tikzpicture}%
        \label{fig:similar_residential_2}
    }
    \hfill
    \subfloat[Residential]{%
        \begin{tikzpicture}
            \node[draw=green, line width=1pt, rounded corners] {
                \includegraphics[width=0.20\columnwidth]{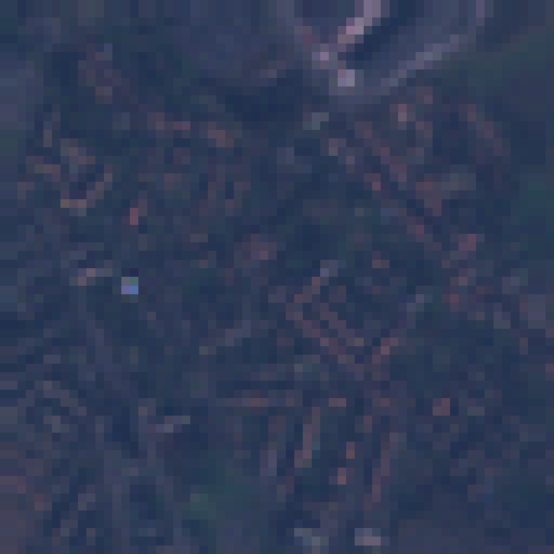}
            };
        \end{tikzpicture}%
            \label{fig:similar_residential_3}
        } \\

    \caption{The figure illustrates instances where models had difficulty in identifying the correct class for remote sensing imagery. The first two rows represent misclassified images: (a)-(d) show permanent crop areas labeled incorrectly, while (e)-(h) show pasture images that were also misclassified. The next two rows (i)-(l) and (m)-(p) display correctly classified images by PIMC for herbaceous vegetation and residential, though these were often confused with other classes by the other methods.}
    \label{fig:euroSAT_examples}
\end{figure}

In the case of the land cover classification task, we observed in the quantitative results of accuracy that all the models did not present a perfect fit in the prediction of the classes. Figure~\ref{fig:euroSAT_examples} illustrates some cases for which the models failed to assign the correct class to the RSI (red boxes) and examples for which only PIMC presented a correct prediction for the image (green boxes). 

A comparison of the results reveals a pattern in the behavior of the mode in the land cover classification tasks. In some cases, samples of RSI belonging to different classes can be difficult to distinguish, even for well-performing models. Figure~\ref{fig:euroSAT_examples} illustrates this issue in the images in the first two rows, where the images belonging to the classes Permanent Crop and Pasture are highly similar in terms of color distribution and shape of the objects (see for example, Figures~\ref{fig:similar_permanentCrop_0} and~\ref{fig:similar_pasture_1}). For this challenging scenario, all models (ImageNet, ViT, SeCo, and PIMC) predicted wrong labels for all samples.

As can be observed in Table~\ref{tab:eurosat_results}, the PIMC model achieved superior performance compared to other models. This improvement highlights the effectiveness of our method in classifying challenging samples. The advantage likely stems from incorporating the vegetation index in multimodal training using two-dimensional representations, enabling better differentiation between classes in vegetation-dense images, which were often misclassified by ImageNet, ViT, and SeCo. 
Images in the second, third, and fourth rows of Figure~\ref{fig:euroSAT_examples} illustrate this scenario. For example, images in Figure~\ref{fig:similar_herbacious_1} and Figure~\ref{fig:similar_herbacious_2} belong to the class Herbaceous Vegetation but are very similar to samples of the Pasture class (e.g., Figure~\ref{fig:similar_pasture_1}) and of the class Residential (e.g., Figure~\ref{fig:similar_residential_0}). In those cases, the models ImageNet and SeCo failed to classify correctly the samples from the last two rows.

\subsubsection{Feature Space}

In addition to the assessment of encoders in downstream tasks in Earth Observation (EO), we analyzed the separation of generated features in the feature space. This analysis provides qualitative insights into how effectively the encoders $\mathbf{I}$ and $\mathbf{T}$ can differentiate features in data that were not considered in their training.

\begin{figure*}[!ht]
\centering
\subfloat[PIMC$^{\text{FT}}$]{\includegraphics[width=.28\columnwidth,height=0.28\textheight,keepaspectratio]{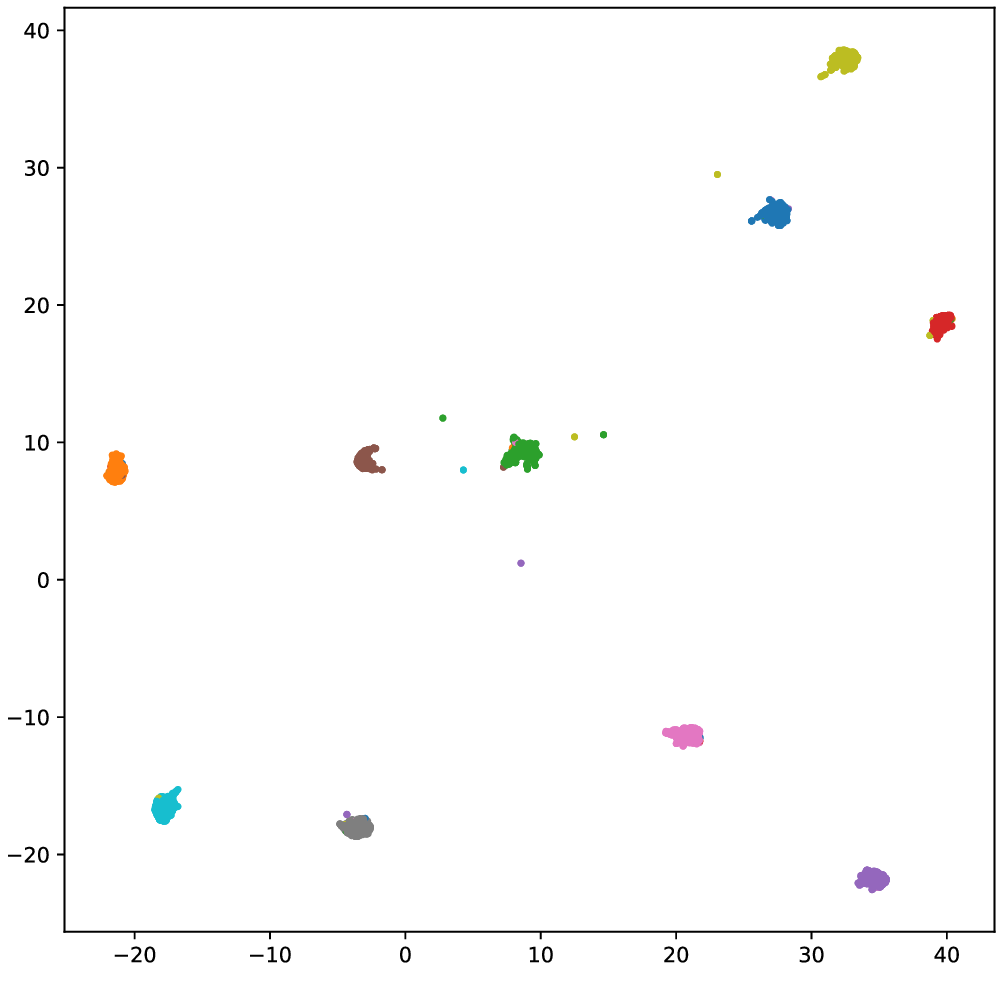}} \hspace{0.2mm}
\subfloat[PIMC]{\includegraphics[width=.28\columnwidth,height=0.28\textheight,keepaspectratio]{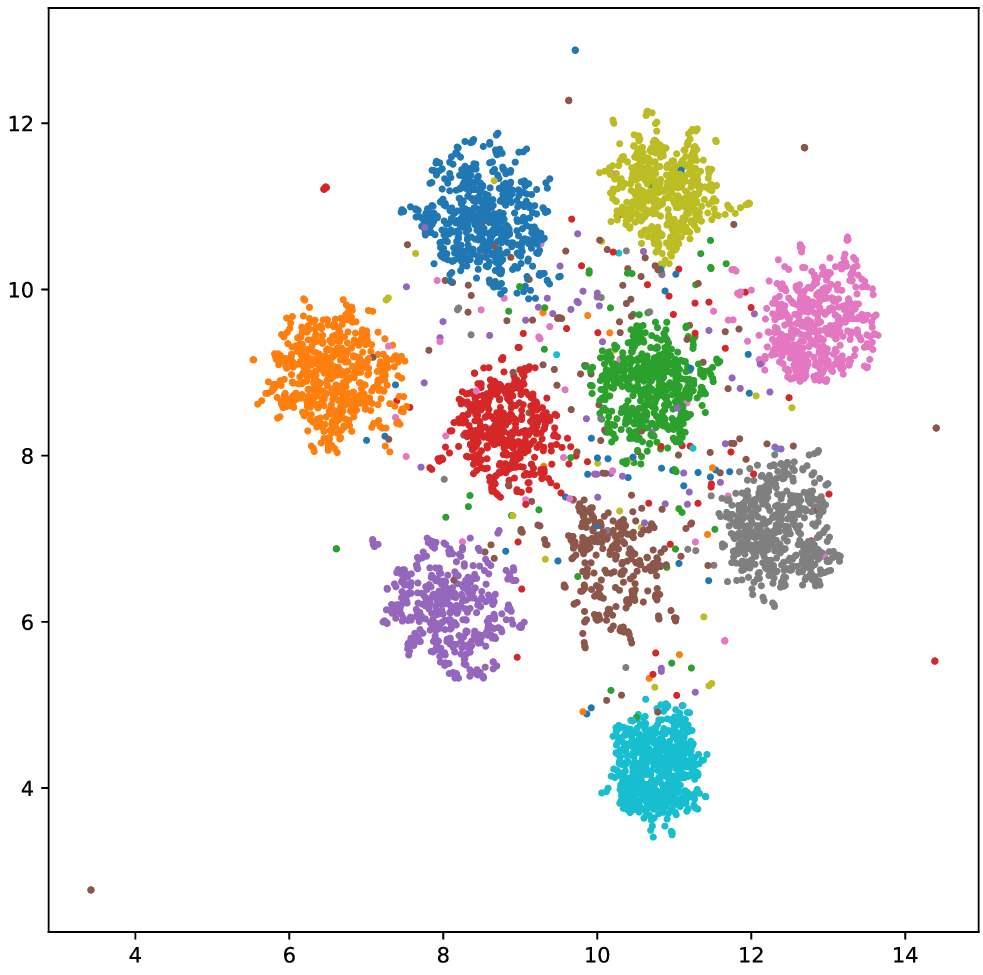}} \hspace{0.15mm}
\subfloat[ImageNet]{\includegraphics[width=.33\columnwidth,height=0.28\textheight,keepaspectratio]{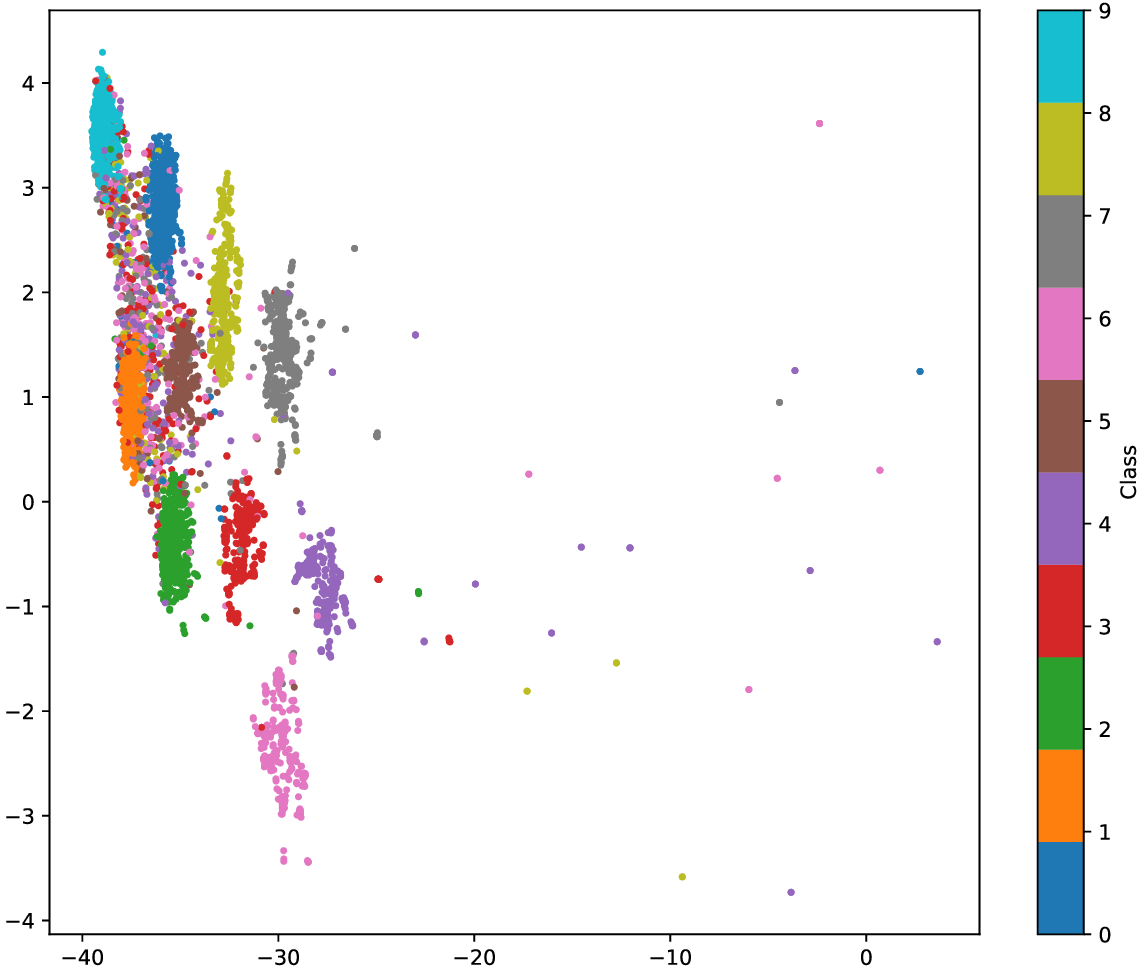}}
\caption{Features from the three encoders applied to the EuroSAT dataset: (a) features from the fine-tuned PIMC model, (b) features from the pre-trained PIMC model, and (c) features from the ImageNet weights without fine-tuning.}
\label{fig:umap_features}
\end{figure*}

Figure~\ref{fig:umap_features} illustrates the features extracted from the last convolutional layer of the ResNet-18 encoders, considering the test set of the EuroSAT dataset. The UMAP technique~\citep{mcinnes2018umap} was applied to create a 2D visualization of these features.

In Figure~\ref{fig:umap_features}(a), the fine-tuned PIMC model for the EuroSAT dataset demonstrates a well-defined feature space, as expected given the labeled nature of the dataset. However, Figure~\ref{fig:umap_features}(b) shows that the features from the PIMC model also exhibit a strong representation, effectively separating and clustering images of the same classes, outperforming the ImageNet pre-training as depicted in Figure~\ref{fig:umap_features}(c). The training of the $\mathcal{I}$ encoder utilized vegetation indices, while the EuroSAT dataset includes classes that are not directly related to crop types (e.g., highways, lakes, and residential areas). This indicates that PIMC can discern attributes in images exhibiting temporal changes, even when these changes are not directly relevant to crop fields or vegetation.

\subsection{Ablation}

\paragraph{{\bf Number of Pixels Impact}}

In our formulation, we tested various numbers of pixels selected from the Peano curve to determine the optimal amount that yields the best performance without requiring excessive computational power. Table~\ref{tab:pixels_pastis_eurosat} presents the results for the downstream task of classification in both time series and the RSI modal.

\begin{table*}[!t]
    \centering
    \caption{Table with the values for the downstream tasks using different encoders trained with different numbers of pixels.}
    \begin{tabular}{@{}ccccccc@{}}
    \toprule
    \textbf{Pixels} & \multicolumn{4}{c}{\textbf{PASTIS}} & \multicolumn{2}{c}{\textbf{EUROSAT}} \\ 
    \cmidrule(lr){2-3} \cmidrule(lr){4-5}
     & \textbf{ACC} & \textbf{BAL ACC} & \textbf{MAE} & \textbf{MSE} & \textbf{ACC} & \textbf{BAL ACC} \\ 
    \midrule
    \textbf{25}  & 32.47 & 25.48 & 0.5264 & 0.3679 & 42.68 & 43.18 \\
    \textbf{50}  & 31.84 & 24.19 & 0.5274 & 0.3704 & 46.85 & 44.95 \\
    \textbf{100} & 32.95 & 24.08 & 0.5194 & 0.3637 & 52.14 & 51.56 \\
    \textbf{150} & 34.73 & 25.91 & 0.5186 & 0.3621 & 53.01 & 52.00 \\
    \bottomrule
    \end{tabular}
    \label{tab:pixels_pastis_eurosat}
\end{table*}

The results indicate a discernible pattern: as the number of pixels increases, the quality of the results in downstream tasks concomitantly improves. This suggests that the model's capacity to produce features is proportional to the number of samples (pixels) from the same regions.

\paragraph{{\bf Time Series Length}}

To define how many values will be used in the time series and its two-dimensional representation, we conducted experiments to check the impact of these time series. Table~\ref{tab:ts_size_pastis_eurosat} demonstrates the impact in accuracy for the time series-based pixel classification and land cover classification tasks using the PIMC model trained with 200 pixels.

\begin{table*}[t!]
    \centering
    \caption{Table with the values of accuracy (ACC) and balanced accuracy (BAL ACC) for the downstream task of classification using different encoders trained with different sizes of time series in the one and two-dimensional representations.}
    \begin{tabular}{@{}ccccccc@{}}
    \toprule
    \textbf{Length} & \multicolumn{4 }{c}{\textbf{PASTIS}} & \multicolumn{2}{c}{\textbf{EUROSAT}} \\ 
    \cmidrule(lr){2-3} \cmidrule(lr){4-5}
     & \textbf{ACC} & \textbf{BAL ACC} & \textbf{MAE} & \textbf{MSE} & \textbf{ACC} & \textbf{BAL ACC} \\ 
    \midrule
    \textbf{25}  & 22.13 & 20.47 & 0.5289 & 0.3633 & 72.13 & 68.47\\ 
    \textbf{50}  & 24.55 & 22.32 & 0.5413 & 0.3828 & 82.34 & 74.66\\ 
    \bottomrule
    \end{tabular}
    \label{tab:ts_size_pastis_eurosat}
\end{table*}

A similar behavior can be observed in the number of pixels related to the length of the time series used as input for training downstream tasks. When the model receives more information from a longer time series, the results improve.

\section{Conclusions}

In this paper, we introduced PIMC, a novel contrastive self-supervised learning method for creating encoders based on multimodal data, including satellite images and pixel-wise two-dimensional time series representations. We described how this framework can be applied, including how to generate two-dimensional representations for time series data and how to tailor trained encoders for different downstream tasks. This comprehensive development addresses the first research question (RQ1).
Based on the results from the models utilizing the 2D representation, we conclude that representing the MSRSI with vegetation index time series, rather than absolute pixel values, effectively produces representative latent spaces, thereby addressing our second research question (RQ2). Furthermore, considering the performance of models trained with 1D representations, we positively respond to the third research question (RQ3), as encoders trained based on the use of 2D representations yielded superior or comparable results across all downstream tasks. However, the PIMC, even after the fine-tuning process, keeps showing difficulty in distinguishing samples from different classes (as demonstrated in Figure~\ref{fig:euroSAT_examples} for the land cover classification task,) especially for imbalanced datasets.

%Based on the results from the pixel time series classification and time series and in image classification we can affirmatively answer our third research question (Q3), as the accuracy achieved with our approach surpasses that obtained using alternative pre-trained weights, such as ImageNet and SeCo.

The study conducted opens up new opportunities for further investigations. For example, researchers could explore different methods for creating two-dimensional representations of time series data, rather than relying solely on recurrence plots. Some alternatives include the use of the Gramian Angular Summation Field (GASF) and the Markov Transition Field (MTF)~\citep{wang2015encoding}. 
Furthermore, further analyses could be conducted on the potential of PIMC-based encoders for various downstream tasks, such as time series anomaly detection, which includes addressing missing data points or occlusions caused by clouds. Other promising research directions involve investigating alternative remote sensing indices and employing more robust networks, such as transformer-based methods, as primary encoders.

\section*{Acknowledgments}

We would like to thank CAPES, CNPq (304836/2022-2), and FAPESP (2022/12294-8 and 2023/11556-1) for their financial support. In addition, this work used resources of the ``Centro Nacional de Processamento de Alto Desempenho em São Paulo'' (CENAPAD-SP).

% References
{
	\begin{spacing}{1.17}
		\normalsize
		\bibliography{references} % Include your own bibliography (*.bib), style is given in isprs.cls
	\end{spacing}
}
\end{document}